\documentclass[10pt]{article} 
\usepackage[accepted]{tmlr}

\usepackage{amsmath,amsfonts,bm}

\def\eqref#1{equation~\ref{#1}}

\def\1{\bm{1}}

\DeclareMathAlphabet{\mathsfit}{\encodingdefault}{\sfdefault}{m}{sl}
\SetMathAlphabet{\mathsfit}{bold}{\encodingdefault}{\sfdefault}{bx}{n}

\usepackage{hyperref}
\usepackage{url}
\usepackage{graphicx}
\usepackage{bbm}
\usepackage{multirow}
\usepackage{subfigure}
\usepackage[normalem]{ulem}
\let\classAND\AND
\let\AND\relax
\usepackage{algorithmic}

\let\AND\classAND
\AtBeginEnvironment{algorithmic}{\let\AND\algoAND}
\usepackage{epstopdf}
\usepackage[ruled, linesnumbered]{algorithm2e}
\usepackage{booktabs}

\title{Learning Federated Neural Graph Databases for Answering Complex Queries from Distributed Knowledge Graphs}

\author{\name Qi Hu \email qhuaf@connect.ust.hk \\
      \addr The Hong Kong University of Science and Technology
      \AND
      \name Weifeng Jiang \email  weifeng001@e.ntu.edu.sg \\
      \addr Nanyang Technological University
      \AND
      \name Haoran Li \email hlibt@connect.ust.hk\\
      \addr The Hong Kong University of Science and Technology
      \AND
      \name Zihao Wang \email zwanggc@connect.ust.hk\\
      \addr The Hong Kong University of Science and Technology
      \AND
      \name Jiaxin Bai \email jbai@connect.ust.hk\\
      \addr The Hong Kong University of Science and Technology
      \AND
      \name Qianren Mao \email maoqr@zgclab.edu.cn\\
      \addr Zhongguancun Laboratory
      \AND
      \name Yangqiu Song \email yqsong@cse.ust.hk\\
      \addr The Hong Kong University of Science and Technology
      \AND
      \name Lixin Fan \email fanlixin@webank.com\\
      \addr WeBank
      \AND
      \name Jianxin Li \email lijx@act.buaa.edu.cn\\
      \addr Beihang University}

\newtheorem{definition}{Definition}[section]

\def\month{07}  
\def\year{2025} 
\def\openreview{\url{https://openreview.net/forum?id=3K1LRetR6Y}} 

\begin{document}

\maketitle

\begin{abstract}
The increasing demand for deep learning-based foundation models has highlighted the importance of efficient data retrieval mechanisms. Neural graph databases (NGDBs) offer a compelling solution, leveraging neural spaces to store and query graph-structured data, thereby enabling LLMs to access precise and contextually relevant information.  However, current NGDBs are constrained to single-graph operation, limiting their capacity to reason across multiple, distributed graphs. Furthermore, the lack of support for multi-source graph data in existing NGDBs hinders their ability to capture the complexity and diversity of real-world data. In many applications, data is distributed across multiple sources, and the ability to reason across these sources is crucial for making informed decisions. This limitation is particularly problematic when dealing with sensitive graph data, as directly sharing and aggregating such data poses significant privacy risks. As a result, many applications that rely on NGDBs are forced to choose between compromising data privacy or sacrificing the ability to reason across multiple graphs. To address these limitations, we propose to learn \underline{Fed}erated \underline{N}eural \underline{G}raph \underline{D}ata\underline{B}ase (FedNGDB), a pioneering systematic framework that empowers privacy-preserving reasoning over multi-source graph data. FedNGDB leverages federated learning to collaboratively learn graph representations across multiple sources, enriching relationships between entities, and improving the overall quality of graph data. Unlike existing methods, FedNGDB can handle complex graph structures and relationships, making it suitable for various downstream tasks. We evaluate FedNGDBs on three real-world datasets, demonstrating its effectiveness in retrieving relevant information from multi-source graph data while keeping sensitive information secure on local devices. Our results show that FedNGDBs can efficiently retrieve answers to cross-graph queries, making it a promising approach for LLMs and other applications that rely on efficient data retrieval mechanisms. Our code is available at \url{https://github.com/HKUST-KnowComp/FedNGDB}.
\end{abstract}

\section{Introduction}
Graph Databases (GDBs) excel at efficiently storing and managing highly interconnected data, leveraging their graph structure to efficiently manage complex relationships.
This capability makes them indispensable for applications such as recommendation systems~\citep{wang2019kgat,cao2019unifying} and fraud detection~\citep{prusti2021credit,sadowski2014fraud}, where the graph structure is critical. 
GDBs offer dynamic data models that adapt to evolving structures and deliver scalable performance for intricate queries.
In the era of deep learning-based foundation models, such as large language models (LLMs), their importance has surged with the advent of Retrieval Augmented Generation (RAG), where agents utilize external GDBs, such as knowledge graphs (KGs) to enhance their retrieval capabilities~\citep{gao2023retrieval, lewis2020retrieval}. 
This integration facilitates the creation of interactive natural language interfaces tailored to domain-specific applications, enabling more intuitive and accessible interaction with structured data and unlocking new possibilities for intelligent data-driven solutions~\citep{matsumoto2024kragen, hussien2025rag, jin2024health}.

\begin{figure}[tbp]
  \centering
  \includegraphics[width=0.8\linewidth]{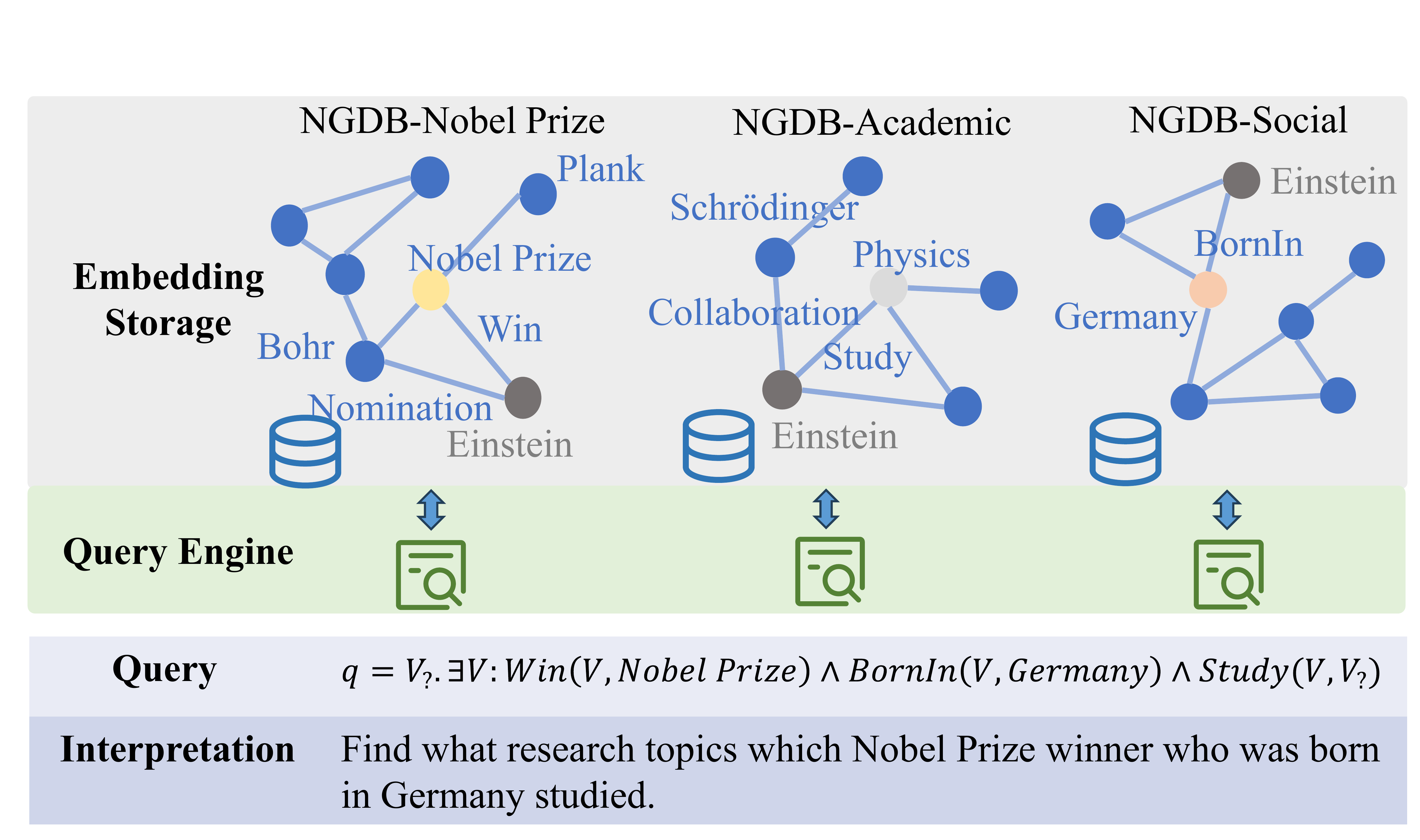}
  \caption{An example of cross graph queries on distributed neural graph databases. The relations and entities in a KG complex query can be from NGDBs which cannot be solved in a single local database.}
  \label{fig:intro}
\end{figure}

However, traditional graph databases often suffer from two limitations: the ineffectiveness of free-text semantic search and graph incompleteness, which are prevalent issues in real-world knowledge graphs and other graph-structured data.
Incompleteness leads to the exclusion of relevant results, as the graph database may not capture all the necessary relationships and connections between entities by traversing~\citep{bordes2013translating, galarraga2013amie}. These limitations hinder their ability to fully support advanced retrieval tasks. 
To address this, neural graph databases (NGDBs) have recently been proposed by~\citet{besta2022neural,ren2023neural}.
They integrate the adaptable structure of graph data models with the powerful processing capabilities of neural networks, allowing efficient storage, effective graph-structured data analysis, and flexible attribute representation~\citep{zhang2024text}.
NGDBs provide unified storage for diverse entries in an embedding space and a neural query engine searching for answers to complex queries from the unified storage~\citep{ren2023neural}. 
These databases unlock stronger capabilities for intelligent data exploration, allowing users to craft complex queries and make informed inferences with the help of advanced neural network techniques~\citep{bai2025top}.
Among these applications, complex query answering (CQA) is an important yet challenging task in graph reasoning and can be used to support various downstream tasks~\citep{baisequential, ren2023neural}. 
CQA aims to retrieve answers that satisfy given logical expressions~\citep{hamilton2018embedding, ren2020query2box}, which are often defined in predicate logic forms with relation projection operations, existential quantifiers $\exists$, logical conjunctions $\wedge$, disjunctions $\vee$, etc. 
As shown in Figure \ref{fig:intro}, given a logical query $q$, our aim is to find all the research topic entities $V_?$ for which there exist Nobel Prize winners $V$ who were born in Germany and conducted studies in that specific field. 

Although neural graph databases have achieved remarkable success in addressing complex query answering tasks, they are limited to utilizing a single central graph database and cannot be extended to multiple databases. 
As data plays an increasingly vital role, NGDBs have experienced rapid growth in scale and scope, aggregating knowledge from diverse domains.
Consequently, constructing a graph database that includes all related entities and relations has become difficult and it is impractical to access a central database with all the data needed~\citep{peng2021differentially,chen2021fede, zhang2022efficient}. 
Collaborations between various NDGB holders are essential for answering more complicated queries. 
For example, as shown in Figure \ref{fig:intro}, there are multiple NGDBs with different domain knowledge. 
A complex query may consist of entities and relations from multiple NGDBs, preventing a local query answering model on a single database from answering those cross-graph queries. 
However, there are various reasons hindering the data sharing between NGDB holders. For example, the growing attention on privacy, regulations such as the General Data Protection Regulation (GDPR), and commercial interest between data holders, etc. Distributed database approaches exist~\citep{nadal2021graph}, but they fail to leverage NGDBs’ neural capabilities fully.

Federated learning offers a potential solution, enabling collaborative model training across distributed participants without sharing raw data~\citep{mcmahan2017communication, yang2019federated}.
In federated knowledge graph embeddings, raw graph triplets are retained on local devices, participants train local models, and only gradients are transferred to learn a global graph embedding model~\citep{chen2021fede, zhang2022efficient, peng2021differentially} under the protection of privacy preservation methods, such as homomorphic encryption (HE)~\citep{paillier1999public}, secure multi-party computation (SMPC)~\citep{mohassel2017secureml}, and differential privacy~\citep{dwork2008differential, geyer2017differentially}. While federated learning has been widely applied in learning graph embeddings, recent studies have indicated that learned representations can still potentially leak privacy~\citep{hu2023independent, duddu2020quantifying}, where an attacker can infer sensitive information from the learned embeddings, besides, existing works only focus on learning high-quality representations for simple downstream tasks, such as knowledge graph completion~\citep{chen2021fede, zhang2022efficient, huang2022fedcke, tang2023fedmkgc}, and lacks the ability to reason over graphs and retrieving answers to complex queries.

To overcome these gaps, we propose to learn \underline{Fed}erated \underline{N}eural \underline{G}raph \underline{D}ata\underline{B}ase (FedNGDB), a novel system that can reason over multi-source graphs avoiding sensitive raw data transmission and embedding exposure to safeguarding privacy. FedNGDB can be applied to different central complex query answering models. It leverages federated learning to train local query answering models in local NGDBs and align graph embeddings for global queries. Different from other federated graph embedding models, the FedNGDB server not only takes the responsibility of aggregating global models but also decomposing given complex queries to sub-queries to compute the query encoding and retrieve answers from distributed NGDBs under the protection of multi-party computation to avoid global model storage. Meanwhile, to better evaluate distributed NGDB systems' performance, we create a benchmark on three widely used datasets. We evaluate our proposed FedNGDBs on the benchmark, the experiment results show the effectiveness of retrieving answers to complex queries from multi-source graphs.
We summarize our major contributions as follows:
\begin{itemize}
    \item To the best of our knowledge, we are the first to extend federated graph embedding systems to complex query answering tasks, which is critical for graph holders' collaboration.
    \item Based on three public datasets, we propose a benchmark for evaluating the retrieval performance of distributed NGDB systems. The benchmark systematically evaluates the retrieval performance facing cross-graph queries.
    \item We propose FedNGDB, a federated neural graph database system that can retrieve answers of complex queries from distributed NGDBs with privacy preserved, solving the drawbacks of existing federated knowledge graph embeddings methods. Extensive experiments on three datasets demonstrate its high performance when facing cross-graph queries.
\end{itemize}

The rest of the paper is organized as follows. We review the related works in Section \ref{sec:related}. Section \ref{sec:preliminary} introduces the preliminary and definition of the distributed complex query answering systems. Section \ref{sec:method} introduces the framework of FedNGDB in detail. Section \ref{sec:exp} evaluates the performance of FedNGDB on the benchmark based on three real-world datasets. Finally, we conclude our work in Section \ref{sec:conclude}.  

\section{Related Work \label{sec:related}}
\subsection{Neural Graph Database}
Neural Graph Databases (NGDBs) neutralize traditional GDBs' storage and query planning modules, aiming for stronger intelligent data exploration capabilities.
Complex query answering (CQA) is a crucial task in NGDBs, which involves training a model to process and answer complex logical queries based on graph reasoning, a process known as query encoding.
These methods represent complex queries into various structures and effectively search for answers among candidate knowledge graph entities. 
GQE~\citep{hamilton2018embedding}, Q2B~\citep{ren2020query2box}, and  HypeE~\citep{liu2021neural} encode queries to vectors, hyper-rectangle and hyperbolic embeddings, respectively. To support negation operators, various encoding methods are proposed. Q2P~\citep{bai2022query2particles} and ConE~\citep{zhang2021cone} use multiple vectors to represent complex queries. BetaE~\citep{ren2020beta}, GammaE~\citep{yang2022gammae}, PERM~\citep{choudhary2021probabilistic} propose to use various probabilistic distributions to encode complex logic graph queries.
Some methods, like LMPNN~\citep{wang2022logical} CQD~\citep{ arakelyan2020complex}, Var2Vec~\citep{wang2023efficient} take pre-train embeddings on simple link prediction tasks and apply logic operators to answer complex queries.
Meanwhile, neural structures are utilized to encode complex queries: BiQE~\citep{kotnis2021answering} and KgTransformer~\citep{liu2022mask} are proposed to use transformers, SQE~\citep{baisequential} applies sequential encoders, GNN-QE~\citep{zhu2022neural} and StarQE~\citep{alivanistos2021query} use message-passing graph neural networks to encode queries, respectively. There are also encoding methods proposed to encode various types of knowledge graphs:
NRN~\citep{bai2023knowledge} is proposed to encode numerical values, MEQE~\citep{bai2023complex} extends logical queries over events, states, and activities. P-NGDB~\citep{hu2024privacy} first extends the privacy preservation to neural graph databases.

While there are numerous existing complex query answering methods, these methods mainly focus on a single large graph. There are some methods proposed to reason over multi-view and temporal and varying graphs: MORA~\citep{xi2022reasoning} ensembles multi-view knowledge graphs to scale up complex query answering, TTransE~\citep{leblay2018deriving} and TRESCAL~\citep{wang2015knowledge} can be applied on temporal knowledge graphs. However, these methods need raw data transmission and privacy protection is not considered. Conducting privacy-preserving complex query answering on multi-source knowledge graphs is still unexplored. With the growing attention on privacy and data protection, sensitive data cannot circulate freely among data holders, and complex query answering is forced to be conducted collaboratively on multiple knowledge graphs. Our research introduces federated learning to existing complex query answering so that we can apply reasoning over distributed NGDBs without raw data sharing. 

\subsection{Federated Knowledge Graph Embedding}
In recent years, federated learning has emerged as a promising approach to address privacy and scalability concerns in machine learning. It allows data owners to participate in model co-construction without raw data transmission to reduce privacy leakage risks~\citep{mcmahan2017communication, yang2019federated} under the protection of privacy protection techniques such as differential privacy (DP)~\citep{yang2019federated}, homomorphic encryption (HE)~\citep{zhang2020batchcrypt}, secure multi-party computation (SMPC)~\citep{byrd2020differentially}. Various studies have been conducted to explore the potential of federated learning in different domains, such as recommender system~\citep{yang2020federated}, finance~\citep{long2020federated} etc. 
Federated databases are proposed to manage distributed data allowing storing and querying databases with privacy preserved~\citep{bater2017smcql, berger2008federated,10.5555/2519214}. Although some federated graph databases are proposed to manage graph-based data~\citep{nadal2021graph}, the study for the latest neural graph databases are still be ignored.

Federated knowledge graph embedding is another related topic. It tries to represent entities and their semantic relations into embedding spaces. FedE~\citep{chen2021fede} learns knowledge graph embeddings locally and aggregates all local models in a global server for higher representation quality. FedR~\citep{zhang2022efficient} proposes to learn representation with privacy-preserving relation aggregation to avoid privacy leakage risks in entity embedding and reduce communication costs. FedCKE~\citep{huang2022fedcke}, FedMKGC~\citep{tang2023fedmkgc} extend federated learning to learn global representations from different domains and multilingual knowledge graphs. FedEC~\citep{CHEN2022109459} applies contrastive learning to tackle data heterogeneity in knowledge graphs. 
MaKEr~\citep{MaKEr} and MorseE~\citep{10.1145/3477495.3531757} utilize meta-learning to transfer knowledge among knowledge graphs to train graph neural networks for unseen knowledge extrapolation and inductive learning. DP-FLames~\citep{10.1145/3543507.3583450} quantifies the privacy threats and incorporates private selection in federated knowledge embeddings. FLEST~\citep{wang2023federated} decomposes the embedding matrix and enables the sharing of latent representations to reduce the risks of privacy leakage and communication costs, FedM~\citep{hu2022federated} splits the duty of aggregating entities and relations to reduce the risks of graph reconstruction attacks. FKGE~\citep{peng2021differentially} applies differential privacy and avoids the need for a central server.
While existing federated knowledge graph embedding methods are proposed to distributedly learn high-quality representations with privacy preservation, there are some works indicate that the learned embeddings are informative and vulnerable to various privacy attacks~\citep{hu2022learning}, even in federated scenarios~\citep{hu2024user, 10.1145/3543507.3583450}. Besides, these methods all lack the ability to answer complex queries on multi-source knowledge graphs which is critical for more complicated downstream tasks. Our research expands the simple federated knowledge graph embedding to answer complex queries on distributed knowledge graphs.

\section{Preliminary and Problem Formulation \label{sec:preliminary}}
\subsection{Preliminary}
Following the general setting of federated knowledge graph embeddings, we denote a set of graph-structured data from various data owners as $\mathcal{G}=\{g_1, g_2,...,g_N\}$, where $N$ is the total number of graphs. Data owners have their own graph data and cannot access to other's databases. Let $g_k=(\mathcal{V}_k, \mathcal{R}_k, \mathcal{T}_k)$ denotes the $k\text{-th}$ graph in $\mathcal{G}$, where $\mathcal{V}_k$ denotes the set of vertices representing entities in the graph $g_k$, $\mathcal{R}_k$ denotes the set of relations, $\mathcal{T}_k$ denotes the set of triplets. Specifically, $\mathcal{T}_k = \{(v_h,r,v_t)\} \subseteq \mathcal{V}_k \times \mathcal{R}_k \times \mathcal{V}_k$ denotes there is a relation between $v_h$ and $v_t$, where $v_h, v_t \in \mathcal{V}_k$, $r \in \mathcal{R}_k$. We denote $\mathcal{V}=\cup_{k=1}^N\mathcal{V}_k$, $\mathcal{R}=\cup_{k=1}^N\mathcal{R}_k$, 
$\mathcal{T}=\cup_{k=1}^N\mathcal{T}_k$ as the set of vertices, relations, and triplets of all graph data, respectively.
    
\subsection{Complex Logical Query}
The complex logical query is defined in existential positive first-order logic form, consisting of various types of logic expressions like existential quantifiers $\exists$, logic conjunctions $\wedge$, and disjunctions $\vee$. In the logical expression, there is a set of anchor entities $V_a \in \mathcal{V}$ denotes given context, existential quantified variables $V_1, V_2,...,V_k \in \mathcal{V}$, and a unique variable $V_?$ denotes our query target. The complex query intends to find the target answers $V_? \in\mathcal{V}$, such that there are $V_1, \cdots, V_k \in \mathcal{V}$ in the graph-structured data that can satisfy the given logical expressions simultaneously. 
Following the definition in \citet{ren2020query2box}, the complex query expression can be converted to the disjunctive normal form (DNF) in the following:

\begin{equation}
\begin{split}
    q[V_?] &= V_?. V_1,...,V_k:c_1 \lor c_2 \lor ... \lor c_n \\
c_i &= e_{i,1} \land e_{i,2} \land ... \land e_{i,m},
\end{split}
\end{equation}
where $e_{i,j}$ is the atomic logic expression, which can be the triplet $(V, r, V')$ denotes relation $r$ between entities $V$ and $V'$, $c_i$ is the conjunction of several atomic logic expressions $e_{i,j}$. $V, V'$ are either anchor entities or existentially quantified variables.


\begin{figure}[ht]
  \centering
  \includegraphics[width=0.85\textwidth]{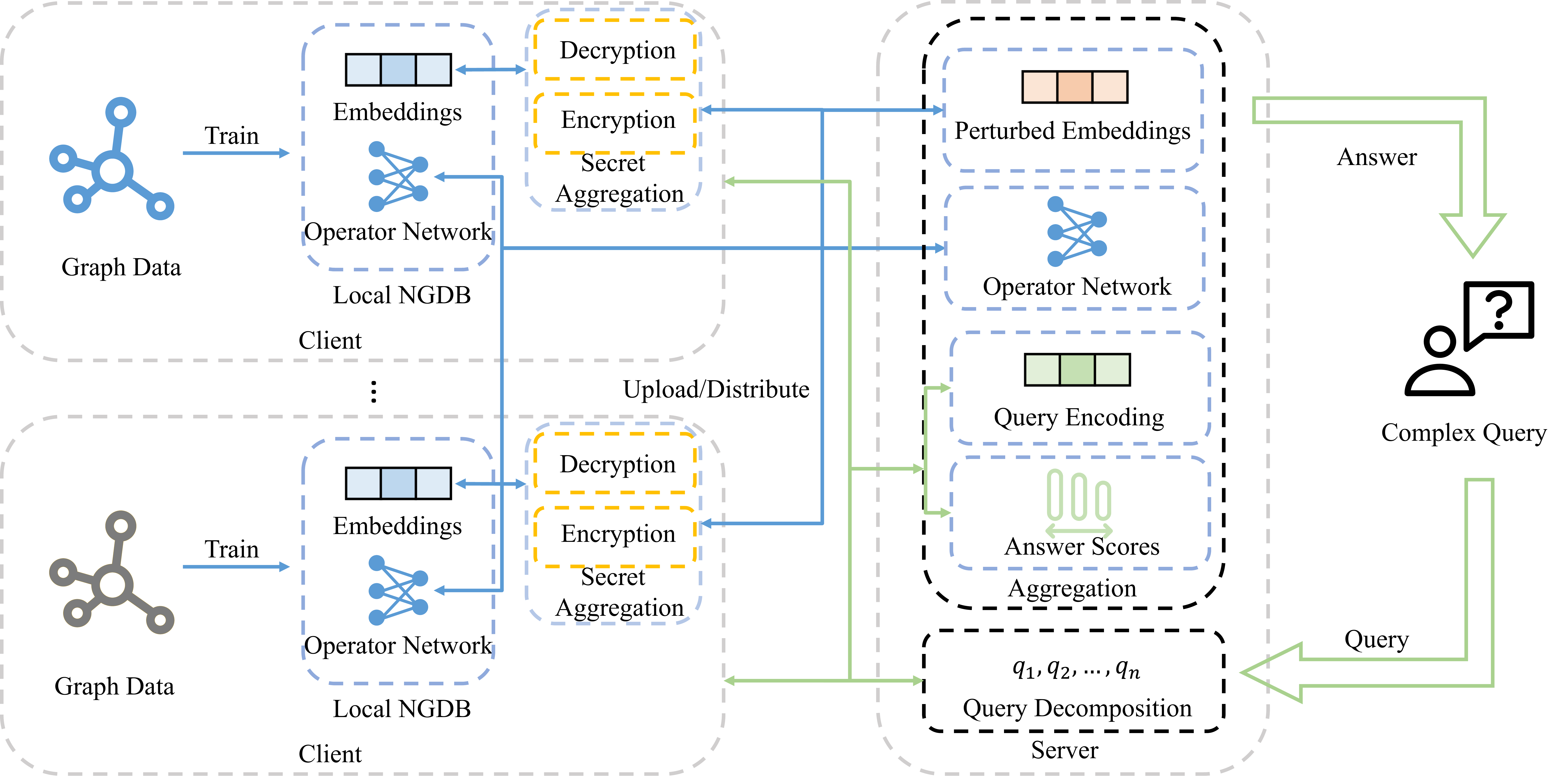}
  \caption{The training and retrieval process of FedNGDB. The blue line denotes the training process, and the green line denotes the retrieval process. In the training, clients' NGDB models are trained on respective graph-structured data. At each round, local NGDBs are aggregated at the server and updated using the global parameters. Among them, the embeddings are protected by secret aggregation so that the server can not access them. In the retrieval, each query is decomposed into sub-queries. Clients compute sub-query embeddings which the server is used to aggregate query embeddings. Answer scores are computed at clients and are aggregated at a server to retrieve answers. 
}
  \label{Fig:framework}
\end{figure}

\subsection{Distributed Graph Set Query}
Graph databases are owned by different data holders and cannot be shared directly with each other, therefore, a complex query $q$ can involve entities and relations from different graphs. We define those queries as follows:
\begin{definition}[Cross-graph Query]
\label{def:cross-graph}
A complex query $q$ is a cross-graph query if there exists query answers $V_? \in \mathcal{V}$, such that there are $V_1, \cdots, V_k \in \mathcal{V}$ in the graph that can satisfy the given logical expressions and the atomic expressions in the query can not be found in a single graph. 
\end{definition}
For example in figure \ref{fig:intro}, the entity "Physics" is the answer to the query $q$ because there exists an existentially quantified variable "Einstein" that can satisfy the logical expression. The query $q$ is a cross-graph query as the atomic expressions in the query are from different graph databases and can not be found in a single graph. For example, Win(Einstein, Nobel Prize) and BornIn(Einstein, Germany) are two atomic expressions from different graph databases. When the distributed graph databases face cross-graph queries, the answer can not be inferred from a single graph. Besides, We have the definition of in-graph query correspondingly:

\begin{definition}[In-graph Query]
\label{def:in-graph}
A complex query $q$ is an in-graph query if for all answers $V_? \in \mathcal{V}$ to the query, such that there are $V_1, \cdots, V_k \in \mathcal{V}$ in the graph that can satisfy the given logical expressions and the atomic expressions in the query are from a single graph database. 
\end{definition}
These in-graph queries can retrieve answers according to one of the graphs in the graph set and can be solved with existing complex query answering models. However, in a distributed graph neural graph database system, the in-graph queries may retrieve more answers as more knowledge is provided.

\subsection{Problem Formulation}
Given a graph-structured data set $\mathcal{G}=\{g_1,...,g_N\}$ with $N$ graphs. Every graph is owned by independent data holders and can not be shared to construct a unified graph database. We assume that the triples are sensitive as they describe the informative relations between entities while the index of entities and relations can be shared, which means that the triples in each graph database will stay private on local devices. The graphs in the $\mathcal{G}$ are related and have part of entities and relation overlapped. There are complex queries involving elements from the graph set $\mathcal{G}$ and can be classified as cross-graph queries and in-graph queries. We aim to construct a distributed neural graph database system to reason over multi-source graphs and retrieve answers to complex logical queries while keeping privacy preserved, especially the triple information indicating the relations between entities. To achieve this, we assume that there is an honest but curious server managing the federated neural graph database system. Because the learned embeddings are vulnerable to various privacy attacks, the embeddings cannot be exposed to the server and should be further protected before being transferred to the server.  

\section{Federated Neural Graph Databases \label{sec:method}}
In this section, we introduce the learning and query retrieval process of our proposed FedNGDB. 

\subsection{Model Learning}
We first introduce the training of FedNGDB. As shown in Figure~\ref{Fig:framework}, FedNGDB has a central server and a set of clients. Each client has a graph with overlapping entities of others. The server takes the responsibility of aggregating parameters and organizing the training and retrieval process. The clients train a local NGDB model based on their graph-structured data. According to the sensitivity of the parameters, we divide the query encoding methods into two parts: operator function with parameter $\Theta$ and entity embeddings $\mathbf{E}$. For the operator function, the client directly sends the parameter to the server for aggregation and receives the global function to update the local operator function, which is similar to FedAvg~\citep{mcmahan2017communication}. Therefore, in the following parts, we only introduce the entity embeddings aggregation in FedNGDB.

\subsubsection{\bf Secret Aggregation}
There are various techniques, like homomorphic encryption (HE)~\citep{paillier1999public}, secure multi-party computation (MPC)~\citep{mohassel2017secureml}, and differential privacy~\citep{dwork2008differential, geyer2017differentially} to protect the uploaded parameters, however the protection of aggregated global model are often ignored. Unfortunately, the global model is informative and vulnerable to privacy attacks~~\citep{zhang2022efficient, hu2024user}. Hence we propose a secret aggregation applied in parameter aggregation which can prevent the global server from knowing the aggregated parameters using homomorphic encryption. Assume that at each client $i$, a parameter denoted as $\theta_i$ is uploaded to a server for aggregating global parameter $\theta$. The procedure of secret aggregation is described in Algorithm~\ref{alg:aggregation}. In the beginning, each client $C_i$ randomly generates perturbed parameters $\theta_i^r$ and shares them with other clients with encryption (Shown in Appendix~\ref{appendix:sharing}). After the sharing, each client has a set of parameters  $\{\theta_{1}^r,\theta_{2}^r.\cdots,\theta_{n}^r\}$. In the training process, for each client, $C_i$ uploads perturbed parameter $(\theta_i+\theta_i^r)$ to the server under the protection of homomorphic encryption. The server collects all perturbed parameters from clients, computes aggregated perturbed parameter $\theta^r$, and sends it back to all clients. Finally, after receiving the perturbed parameters, the clients first decrypt the parameters and can compute aggregated parameter $\theta$ by removing the perturbed parameters and preventing exposing it to the server. 

Besides, the parameters are also protected by the differential privacy to prevent the potential privacy leakage~\citep{zhu2019deep}, we use local differential privacy (LDP) to the local gradients to further improve the difficulty to recover sensitive information from transmission. As proposed in~\citet{yi2021efficient}, we clip the local gradients based on the $L_\infty$ norm with the threshold $\mathcal{C}$ and apply zero-mean Laplacian noise with strength $\lambda$, the upper bound of the privacy budget $\epsilon$ is $\frac{2\mathcal{C}}{\lambda}$. The detailed discussion can be found in Appendix \ref{sec:DP}.

\subsubsection{\bf Model Training}
Similar to~\citet{chen2021fede}, the server constructs a set of mapping matrices $\{\mathbf{M}^i\ \in \{0,1\}^{n \times n_i}\}_{i=1}^{N}$ and existence vectors $\{\mathbf{v}^i \in \{0,1\}^{n \times 1} \}_{i=1}^{N}$ to denote the entities in each client, where $n$ is the number of all unique entities in KG set and $n_i$ is the number of entities from client $C_i$. $\mathbf{M}^i_{m,n} = 1$ if the $m$-th entity in entity table corresponds to the $n$-th entity from client $C_i$.  $\mathbf{v}^{i}_m = 1$ indicates that the $m$-th entity in entity table exists in client $C_i$.

FedNGDB performs secret aggregation for entity embeddings. First, the client $C_i$ will randomly initialize the local entity embeddings $\mathbf{E}_0^i \in \mathbb{R}^{n_i\times d}$ and perturbation embeddings $\mathbf{E}_r^i \in \mathbb{R}^{n_i\times d}$. Every client will share the permutation embeddings with all clients. At round $t$, the server will select part of the clients $\mathbf{C}$ participating in the training. After local training of CQA on respective local graphs, client $C_i$ sends perturbed local entity embeddings $\mathbf{E}_t^i+\mathbf{E}_r^i$ to the server. The server will aggregate the entity embeddings using perturbed local embeddings:
\begin{equation}
\begin{aligned}
	\mathbf{E}_{t+1}^r \leftarrow \left( \mathbbm{1} \oslash \sum_{i \in \mathbf{C}} \mathbf{v}^{i} \right) \otimes \sum_{i \in \mathbf{C}}\mathbf{M}^{i} (\mathbf{E}_{t+1}^i+\mathbf{E}_r^i),
\label{eq:ent-avg}
\end{aligned}
\end{equation}
where $\mathbbm{1}$ denotes all-one vector, $\oslash$ denotes element-wise division for vectors and $\otimes$ denotes element-wise multiply with broadcasting. After aggregation, the server sends the aggregated entity embeddings back to all clients, and the client $C_i$ receives:

\begin{equation}
\begin{aligned}
	\mathbf{E}_{t+1}^{r,i} \leftarrow {\mathbf{M}^i}^\top \mathbf{E}_{t+1}^{r},
\label{eq:ent-1}
\end{aligned}
\end{equation}
and the client $C_i$ can compute and update the local entity embeddings as:

\begin{equation}
\begin{aligned}
	\mathbf{E}_{t+1}^{i} \leftarrow  \mathbf{E}_{t+1}^{r,i} 
 - {\mathbf{M}^i}^\top \left( \mathbbm{1} \oslash \sum_{j \in \mathbf{C}} \mathbf{v}^{j} \right) \otimes \sum_{j \in \mathbf{C}}\mathbf{M}^{j} \mathbf{E}^{j}_{r}.
\label{eq:ent-2}
\end{aligned}
\end{equation}
After secret aggregation, the entity embeddings of all clients are shared without exposing the sensitive information to the server. Besides the training of entity embeddings, the operator networks are trained and aggregated using FedAvg~\citep{mcmahan2017communication}, and the detailed descriptions of the FedNGDB are shown in Algorithm \ref{alg:FedNGDB}.

\subsection{Query Retrieval}
After training, the server in FedNGDB manages the process of retrieving answers to complex queries. 
The server first tries to arrange related clients to encode the coming queries and retrieves answers from all local graph databases based on the encoding.

\subsubsection{\bf Query Encoding}
Query encoding methods commonly represent queries to embeddings and retrieve answers according to scoring functions where similarity functions are widely used. 
FedNGDB encodes queries and treats two types of queries differently. 
For in-graph queries, as these queries only involve a single graph data, FedNGDB can directly encode queries using corresponding local complex query answering models.   
For cross-graph queries, as the query involves entities from multiple graphs, the server will take the responsibility to plan the entire encoding process: First, the server will decompose the query to atomic expressions and send each expression to corresponding local graphs. The clients encode received atomic queries using their own local CQA models and send back the results to the server. The server collects all the encoding results and uses global operator function models to compute the representations of the queries. The query is iteratively updated by communicating between the server and clients until the original queries are encoded.

\subsubsection{\bf Answer Retrieval} Because the entity embeddings are not stored in the central server, we can only retrieve answers from all distributed local graphs after encoding the queries. Given a query encoding $q$, we score all the candidate entities at each local graph database, at client $C_i$:

\begin{equation}
\begin{aligned}
	\mathbf{S}^i_q \leftarrow f^i_s(\mathcal{V}_i, q) \in \mathbb{R}^{n_i \times 1},
\label{eq:ent-3}
\end{aligned}
\end{equation}
where $f^i_s$ is a score function in the client $C_i$. Then the score will be uploaded to the server to aggregate a score table for all unique entities in the graph sets:

\begin{equation}
\begin{aligned}
	\mathbf{S} \leftarrow \left( \mathbbm{1} \oslash \sum_{i=1}^N \mathbf{v}^{i} \right) \otimes \sum_{i =1}^N \mathbf{M}^{i}\mathbf{S}^i.
\label{eq:ent-4}
\end{aligned}
\end{equation}
The final answers to the queries are retrieved globally from the graph database set according to the score table.

\section{Experiments \label{sec:exp}}
In this section, we create a benchmark of distributed graph complex logical query answering problems for distributed neural graph databases and evaluate our proposed FedNGDB's performance on the benchmark.

\subsection{Datasets and Experiment Setting}
We introduce the detailed information of our used datasets and the setting of our experiments.

\begin{figure*}[ht]
  \centering
  \includegraphics[width=0.99\textwidth]{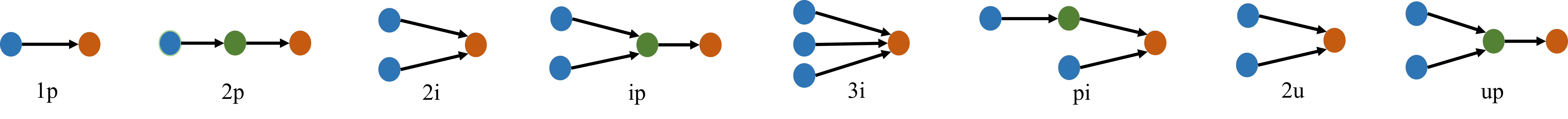}
  \caption{The query structures used for evaluation in the experiments. Naming for each query structure is provided under each subfigure, for brevity, the p, i, and u represent the projection, intersection, and union operations respectively.}
  \label{fig:query_types}
  \vspace{-0.2cm}
\end{figure*}

\subsubsection{\textbf{Datasets}}
In our experiment, following previous work, we use the three commonly used knowledge graphs in central graph reasoning tasks as graph-structured data: FB15k~\citep{bollacker2008freebase, bordes2013translating}, FB15k-237~\citep{toutanova2015observed}, and NELL995~\citep{carlson2010toward} to construct the distributed query answering benchmark, which can make better comparison to existing works. In each dataset, there are vertices describing entities and edges describing relations. To evaluate the distributed complex query, we follow the common settings of federated knowledge graph emebddings~\citep{chen2021fede,zhang2022efficient}, conducting experiments assuming having 3 and 5 clients in each federated neural graph database system, respectively. For more clients, we also conduct simple exeperiments as shown in Appendix~\ref{appx: more clients}. We randomly select relations into clients and split triples into clients according to selected relations as a local graph database. The triplets in each local graph are separated into training, validation, and testing with a ratio of 8:1:1 respectively. Following previous works~\citep{ren2020query2box}, we construct training graph $\mathcal{G}_{train}$, validation graph $\mathcal{G}_{val}$, and test graph $\mathcal{G}_{test}$ in each client by training edges, training+validation edges, and training+validation+testing edges, respectively. The detailed statistics are listed in the table \ref{tab:kg_stats}. \#Clients denotes the number of clients, \#Nodes, \#Relations, \#Edges denote the average number of nodes, relations, and edges in each client respectively.

\begin{table}[t]
\caption{The statistics of three datasets used for experiments. }
\centering
\scalebox{1}{
\label{tab:kg_stats}
\begin{tabular}{ccccc}
\toprule
Graphs & \#Clients & \#Nodes & \#Relations &  \#Edges \\
\midrule
\multirow{2}{*}{FB15k-237} & 3 &13,651  & 79 &103,359\\
 & 5 & 12,639 & 47.4 & 62,015 \\
 \midrule
\multirow{2}{*}{FB15k} & 3 & 14,690 & 448.3&197,404\\
 & 5 & 14,279 & 269& 118,442\\
 \midrule
\multirow{2}{*}{NELL995} & 3 & 40,204 & 66.7& 47,601\\
 & 5 & 28,879 & 40& 28,560\\
\bottomrule
\end{tabular}}
\end{table}  

\subsubsection{\textbf{Query Sampling}}
Following previous work~\citep{hamilton2018embedding, bai2023knowledge}, we evaluate the complex logical query answering performance on the following eight general query types with abbreviations of $1p,2p,2i,ip,$ $3i,pi,2u$, and $up$. As shown in figure \ref{fig:query_types}, each subgraph denotes one query type, where each edge represents either a projection or a logical operator, and each node represents either a set of entities, the anchor entities and relations are to be specified to instantiate logical queries. We use the sampling method commonly used in previous works~\cite {baisequential, ren2020query2box} to randomly sample complex queries from graphs. We randomly sample two sets of queries from the graph sets: in-graph queries and cross-graph queries to evaluate local and global answer retrieval performance. For local model evaluation, we first obtain training, validation, and testing queries from the formerly constructed local graph databases respectively. Then for the training queries, we conduct a graph search to find corresponding training answers on the local training graph. For the validation queries, we search for the answers on both the training graph and the validation graph and only use those queries that have different numbers of answers on the two graphs. For the testing queries, we use those queries that have different answers on the testing graph from answers on the training graph and validation graph. For global model evaluation, we construct global training, validation, and testing graphs using all local graphs, and sample testing queries with atomic expressions from different local graphs, finally we search for answers on three global graphs and only use those queries that have different answers on the testing graph from other two graphs. We collect statistics of complex queries in three datasets and the statistics are shown in Table~\ref{tab:queries_stats}. The number of in-graph queries is the average number of the client's local queries. 

\begin{table}[t]
\caption{The statistics of queries sampled from three datasets used for experiments. }
\centering
\label{tab:queries_stats}
\scalebox{1}{
\begin{tabular}{cccccc}
\toprule
\multirow{2}{*}{Graphs}    & \multirow{2}{*}{\#C} & \multicolumn{3}{c}{In-graph}    & Cross-graph       \\
                           &                            & Train. & Valid. & Test. & Test.   \\
\midrule 
\multirow{2}{*}{FB15k-237} & 3 &317,226  & 11,528 &11,539 &32,573\\
 & 5 & 180,552 & 6,619&6,673&31,469 \\
 \midrule
\multirow{2}{*}{FB15k} & 3 & 592,573 & 19,206&19,267&53,660\\
 & 5 & 344,418 & 11,409& 11,437&53,154\\
 \midrule
\multirow{2}{*}{NELL995} & 3 & 208,070 & 8,810& 8,750&24,954\\
 & 5 & 117,231 &5,177& 5,118&24,237\\
\bottomrule
\end{tabular}}
\end{table}  
 
\begin{table*}[t]

\caption{The retrieval performance of distributed neural graph databases when there are 3 clients. The average results of HR@3 and MRR of all clients are reported. The best results are underlined. The best results of distributed models are in bold.}
\centering
\label{tab:exp-main}
\scalebox{0.8}{
\begin{tabular}{c|c|rrrr|rrrr|rrrr}

\toprule

                            & \multicolumn{1}{c|}{}                          & \multicolumn{4}{c|}{GQE}                                              & \multicolumn{4}{c|}{Q2P}                                              & \multicolumn{4}{c}{Tree-LSTM}                                           \\ \cline{3-14} 
                            & \multicolumn{1}{c|}{}                          & \multicolumn{2}{c}{In-graph}      & \multicolumn{2}{c|}{Cross-graph}  & \multicolumn{2}{c}{In-graph}      & \multicolumn{2}{c|}{Cross-graph}  & \multicolumn{2}{c}{In-graph}       & \multicolumn{2}{c}{Cross-graph}    \\
\multirow{-3}{*}{Graph}     & \multirow{-3}{*}{Setting} & HR@3 & MRR & HR@3 & MRR & HR@3 & MRR & HR@3 & MRR & HR@3 & MRR  & HR@3 & MRR  \\ \midrule
                            & Local  &12.64&12.03& - & - & 14.55 & 13.63 &-  & - &13.32  &12.73 & - &- \\
                            & Central&13.13 & 12.39 & \underline{13.03} & \underline{12.28} & 14.93 & \underline{14.66} & \underline{15.02} & \underline{14.81} & 13.28 & 12.61 & \underline{13.36} & \underline{12.91} \\
                            & FedE &  \underline{\textbf{13.72}} & \underline{\textbf{13.23}} & \textbf{12.74} & \textbf{11.63} & 14.82 & 14.27 & 14.79 & 13.93 & 13.12 & 12.23 & \textbf{12.62} & \textbf{12.08}\\
                            & FedR   &12.89&11.98&- & -&14.32 &14.23&-&- &\underline{\textbf{13.92}} & \underline{\textbf{12.92}}& - &-\\ 
\multirow{-5}{*}{FB15k-237} & FedNGDB  &13.54 & 12.43 & 12.63 & 11.32 & \underline{\textbf{15.32}} & \textbf{14.32} & \textbf{14.83} & \textbf{14.11} & 12.93 & 12.11 & 12.55 & 11.96\\  \midrule
                            & Local  & 22.05 & 18.21&- & - & 24.32 & 22.64 & - &- &22.87 & 20.51& -& -\\
                            & Central & \underline{29.53} & 25.65 & \underline{30.21} & \underline{25.33} & 38.62 & 34.14 & 38.03 & 34.36 & \underline{38.87} & \underline{35.86} & \underline{37.97} & \underline{36.13}\\
                            & FedE  &24.31 & 26.74 & \textbf{27.95} & \textbf{25.21} & 43.68 & \underline{39.62} & 39.72 & 35.95 & 34.27 & 30.18 & \textbf{31.19} & 26.03 \\
                            & FedR  & 20.29 & 18.61 &- &- &25.32  & 22.71& - &-&23.64 &20.97&-&-\\ 
\multirow{-5}{*}{FB15k}     & FedNGDB &\textbf{25.63} & \underline{\textbf{26.87}} & 24.77 & 25.17 & \underline{\textbf{44.02}} & \textbf{39.27} & \underline{\textbf{40.27}} & \underline{\textbf{36.31}} & \textbf{34.85} & \textbf{33.83} & 31.80 & \textbf{28.99} \\  \midrule
                            & Local & 11.85 & 11.03 & - & -& 15.86  &13.02&- &- &13.85 & 13.85&12.94 & - \\
                            & Central &12.87 & 11.95 & 13.06 & 12.46 & 16.74 & 14.82 & \underline{16.42} & 15.63 & 15.41 & 14.23 & \underline{16.27} & \underline{15.83} \\
                            & FedE & 13.29 & 12.72 & 12.46 & 11.82 & \underline{\textbf{17.23}} & 14.12 & \textbf{16.28} & 14.01 & 14.27 & 13.81 & 14.18 & 13.71  \\
                            & FedR &12.01 & 11.23 & - & - & 16.04 & 13.26 & - & - & 12.48 & 11.67 & - & -\\ 
\multirow{-5}{*}{NELL995}   & FedNGDB & \underline{\textbf{14.21}} & \underline{\textbf{13.27}} & \underline{\textbf{13.76}} & \underline{\textbf{12.67}} & 16.62 & \underline{\textbf{15.28}} & 16.27 & \underline{\textbf{16.23}} & \underline{\textbf{16.28}} & \underline{\textbf{15.38}} & \textbf{16.09} & \textbf{15.27}\\ 
 
\bottomrule
\end{tabular}
}
\end{table*}

\subsubsection{\textbf{Baselines}}
We can use various existing query encoding methods as our local base model, to evaluate the effectiveness and generalization ability of our proposed FedNGDB, we select three commonly used complex encoding methods GQE~\citep{hamilton2018embedding}, Q2P~\citep{bai2022query2particles}, Tree-LSTM~\citep{baisequential} as our base model. GQE is a graph query encoding model that encodes a complex query into a vector in embedding space; Q2P represents complex queries using multiple vectors; Tree-LSTM recursively represents complex queries and treats all operations, entities, and relations as tokens.

To the best of our knowledge, there are no existing federated complex query answering methods but several federated knowledge graph embedding methods, therefore, we choose to compare our methods with FedE~\citep{chen2021fede} and FedR~\citep{zhang2022efficient}, two commonly used federated knowledge graph embedding methods as baselines. FedE aggregates both entity embeddings and relation embeddings in a server, while FedR only aggregates relation embeddings for privacy concerns and communication efficiency. We utilize these two methods with slight modifications to train a global complex query answering model: FedE aggregates all query encoding parameters and FedR aggregates relation embeddings and query encoding networks. Besides, we also compare our FedNGDB with local and central settings. These two baselines represent the lower bound and upper bound, respectively, in terms of the amount of information available to the training system. In the local setting, there is no collaboration between clients, while in the central setting, all distributed graphs in the graph set are aggregated for a global graph for training, we sample complex queries from global training, and validation graphs for training and validation. 

If there is no further statement, we use the following implementation settings in the experiments.
We tune hyper-parameters on the validation local queries for the base query encoding methods and set the dimension of entities and relations as 400 for all models for fair comparison and use AdamW~\citep{loshchilov2018decoupled} as optimizer. We set the gradient clip threshold $C=0.1$ and Laplacian noise scale $\lambda=0.2$ to achieve $1$-DP. 

\begin{table*}
\caption{The retrieval performance of distributed neural graph databases when there are 5 clients. The average results of HR@3 and MRR of all clients are reported. The best results are underlined. The best results of distributed models are in bold.}

\centering
\label{tab:exp-client}
\scalebox{0.79}{
\begin{tabular}{c|c|rrrr|rrrr|rrrr}
\toprule

                            & \multicolumn{1}{c|}{}                          & \multicolumn{4}{c|}{GQE}                                              & \multicolumn{4}{c|}{Q2P}            & \multicolumn{4}{c}{Tree-LSTM}    \\ \cline{3-14} 
                            & \multicolumn{1}{c|}{}                          & \multicolumn{2}{c}{In-graph}      & \multicolumn{2}{c|}{Cross-graph}  & \multicolumn{2}{c}{In-graph}      & \multicolumn{2}{c|}{Cross-graph}  & \multicolumn{2}{c}{In-graph}       & \multicolumn{2}{c}{Cross-graph}    \\
\multirow{-3}{*}{Graph}     & \multirow{-3}{*}{Setting} & HR@3 & MRR & HR@3 & MRR & HR@3 & MRR & HR@3 & MRR & HR@3 & MRR  & HR@3 & MRR  \\ \midrule
                            & Local& 11.44 & 10.65  & - & - & 14.65 & 13.8 & - & - & 11.23 & 10.37 & - & - \\
                            &Central&\underline{13.72}&\underline{12.87}&\underline{12.99}&\underline{12.74}&15.93&14.62&\underline{15.74}&\underline{15.23}&\underline{13.62}&\underline{12.54}&\underline{11.86}&\underline{11.53} \\
\multirow{-3}{*}{FB15k-237} & FedNGDB  & \textbf{12.42} & \textbf{11.60} & \textbf{11.20} & \textbf{10.79} & \underline{\textbf{16.13}} & \underline{\textbf{15.78}} & \textbf{15.28} & \textbf{14.91} & \textbf{12.48} & \textbf{11.91} & \textbf{11.49} & \textbf{11.02} \\ 
 \midrule
                            & Local &19.83 & 17.51 & - & - & 36.10 & 35.04 & - & - & 20.03 & 18.62 & - & - \\
                            &Central&20.58&20.14&\underline{21.21}&\underline{21.02}&\underline{42.56}&\underline{41.87}&\underline{39.57}&\underline{39.46}&\underline{24.88}&\underline{24.76}&\underline{24.93}&\underline{24.72} \\
\multirow{-3}{*}{FB15k}    & FedNGDB   & \underline{\textbf{21.40}} & \underline{\textbf{20.83}} & \textbf{20.71} & \textbf{19.94} & \textbf{40.81} & \textbf{37.96} & \textbf{38.56} & \textbf{35.73} & \textbf{24.59} & \textbf{23.75} & \textbf{23.85} & \textbf{22.90} \\
 \midrule
                            & Local  & 10.48 & 10.09 & - & - & 15.26 & 14.37 & - & - & 14.52 & 13.89 & - & - \\
                            &Central&\underline{14.56}&\underline{14.28}&\underline{13.75}&\underline{13.42}&\underline{15.63}&15.14&\underline{15.27}&\underline{14.75}&\underline{16.47}&\underline{15.92}&\underline{15.74}&\underline{14.47} \\ 
\multirow{-3}{*}{NELL995}    & FedNGDB& \textbf{13.79} & \textbf{13.27} & \textbf{12.74} & \textbf{12.18} & \textbf{15.44} & \underline{\textbf{15.81}} & \textbf{15.28} & \textbf{14.24} & \textbf{15.68} & \textbf{14.28} & \textbf{14.57} & \textbf{12.89}\\

\bottomrule
\end{tabular}
}
\end{table*}

\subsubsection{\textbf{Evaluation Metrics}}
Following the previous work~\citep{bai2023knowledge}, we evaluate the generalization capability of models by calculating the rankings of answers that cannot be directly retrieved from an observed graph. Given a testing query $q$,  the training, validation, and public testing answers are denoted as $M_{train}$, $M_{val}$, and $M_{test}$, respectively. We evaluate the quality of retrieved answers using Hit ratio (HR) and Mean reciprocal rank (MRR). HR@K metric evaluates the accuracy of retrieval by measuring the percentage of correct hits among the top K retrieved items. The MRR metric evaluates the performance of a ranking model by computing the average reciprocal rank of the first relevant item in a ranked list of results. The metric can be defined as:
\begin{equation}
   \text{Metric}(q) = \frac{1}{|M_{test}/ M_{val}|} \sum_{v \in M_{test}/M_{val}} m(rank(v)),  \\ \label{equa:metrics}
\end{equation}
 $m(r) = \textbf{1}[r\leq K]$ if the metric is HR@K and  $m(r) = \frac{1}{r}$ if the metric is MRR. Higher values denote better reasoning performance. We train local models at each client by using the in-graph training queries and tune hyper-parameters using the validation queries. The evaluation is then finally conducted on the testing queries, including the evaluation of in-graph queries on local query encoding models and cross-graph queries on the global federated neural graph database system, respectively.

\subsection{Performance Evaluation}
We evaluate FedNGDB's complex query answering performance on three datasets and compare it to other baselines. We apply FedNGDB on three base query encoding models and evaluate the average retrieval performance on various queries. The results are summarized in Table~\ref{tab:exp-main} and Table~\ref{tab:exp-client}. Table~\ref{tab:exp-main} reports the retrieval performance of various distributed graph complex query answering models when there are 3 clients. For each model, we evaluate performance facing in-graph queries and cross-graph queries, respectively. For in-graph queries, the average scores of all clients are reported. We report results in HR@3 and MRR which higher scores indicate better performance. The best results are underlined. The best results of distributed models are in bold. As shown in Table~\ref{tab:exp-main}, our proposed methods can effectively retrieve complex query answers from distributed graph databases. In comparison to local settings, we can see that FedNGDB can utilize all participated local graph databases and performs better facing in-graph queries. For example, GQE model with FedNGDB can achieve 14.21 HR@3 on average while can only reach 11.85 without collaboration. Besides, in comparison to other federated knowledge graph embedding methods, our proposed FedNGDB can reach comparable performance in both in-graph queries and cross-graph queries without exposing sensitive entity embeddings to the server. For example, FedNGDB achieves the best performance in cross-graph queries in more than half of datasets and base query encoding models.

In Table~\ref{tab:exp-client}, we present the performance of FedNGDB when there are 5 clients. We compare the performance with local training without collaboration to demonstrate the influence of client numbers. As shown in the table, FedNGDB performs well compared to complex query answering models without collaboration, there are performance improvements in all datasets after applying FedNGDB to various base query encoding models, demonstrating that FedNGDB can utilize the intrinsic information in the distributed knowledge sets. The collaboration allows FedNGDB to reason over various logical paths to improve performance.

\subsection{Query Types}
\begin{figure*}
\centering
\subfigure[Cross-graph Queries]{\includegraphics[width=0.47\textwidth]{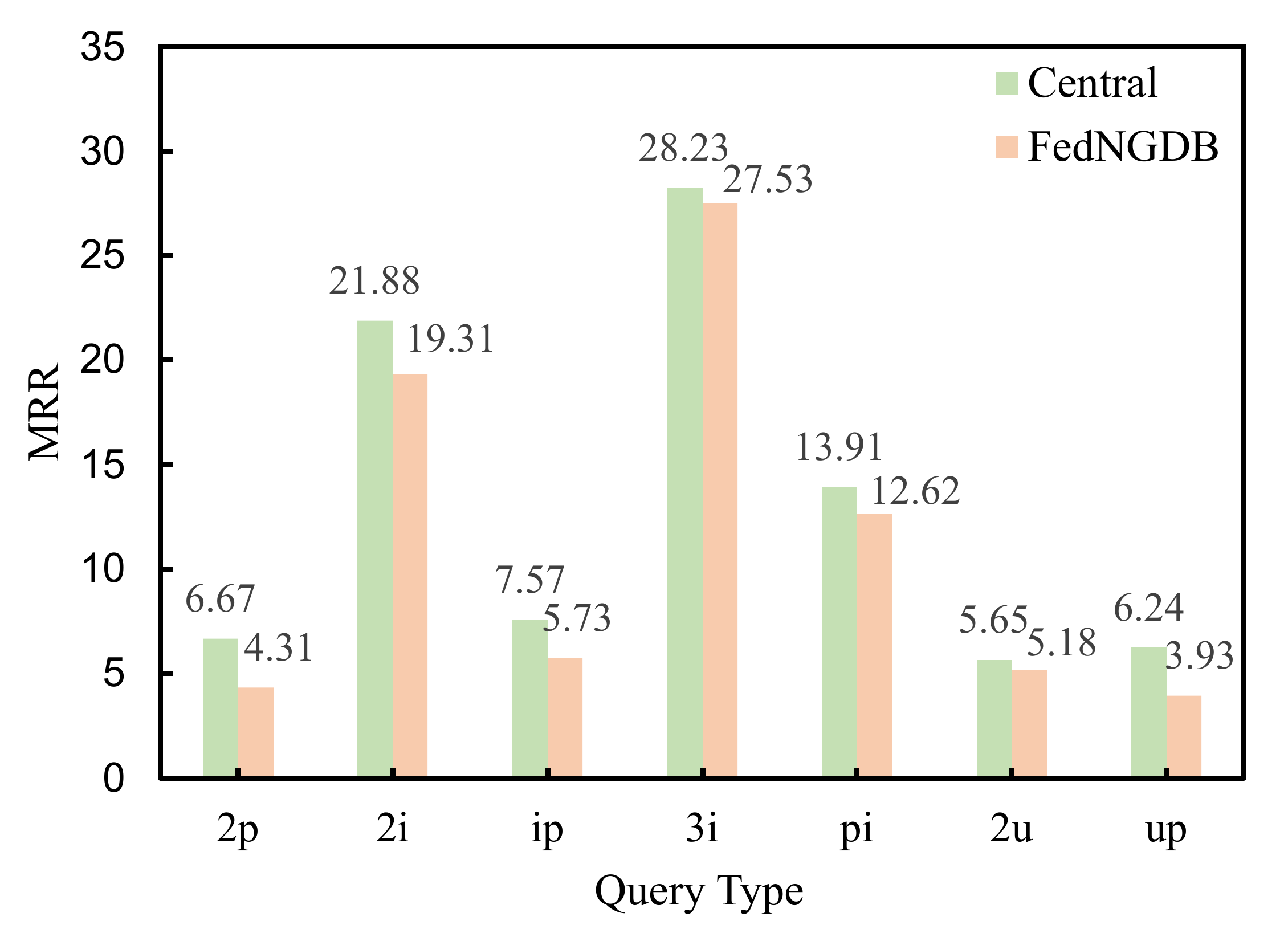}
\label{fig:query_types_exp_1}}\hfil
\subfigure[In-graph Queries]{\includegraphics[width=0.47\textwidth]{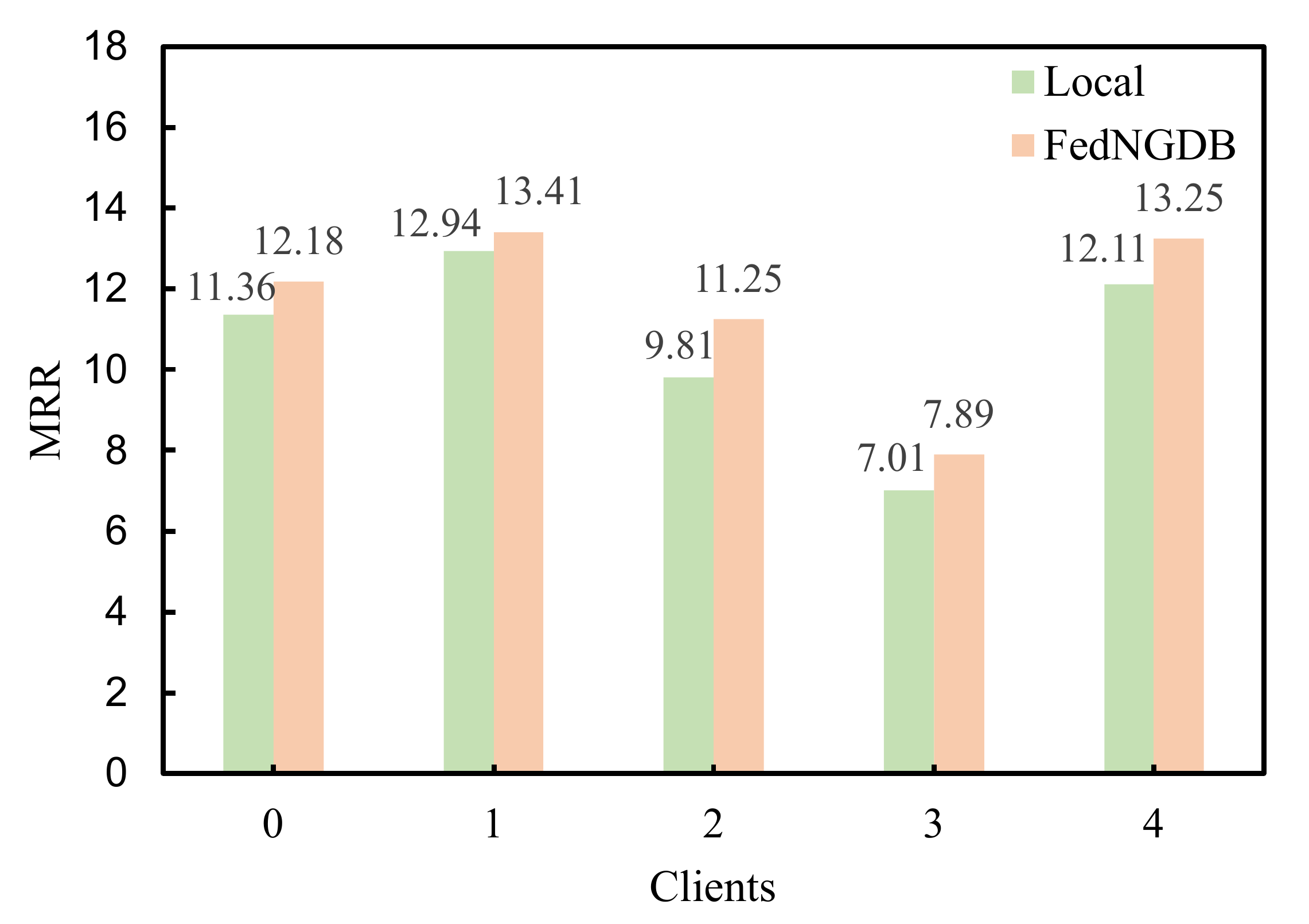}
\label{fig:query_types_exp_2}}
\caption{The evaluation results of FedNGDB-GQE facing cross-graph queries on FB15k-237 (subfigures (a)).  The evaluation results of FedNGDB-GQE facing in-graph queries on FB15k-237 (subfigures (c)).}
\label{distribution_pics}
\end{figure*}

FedNGDB can globally reason over distributed graphs and retrieve answers to cross-graph queries. To evaluate FedNGDB's performance on various types of complex queries, we conduct experiments to evaluate the retrieval performance of FedNGDB and compare it to central learning on FB15k-237 when there are 3 clients. Figure~\ref{fig:query_types_exp_1} shows the FedNGDB-GQE's performance facing cross-graph queries. As we can see from the figure, FedNGDB performs well on most various types of queries compared to the central model. For example, on query types '2i' and 'pi', FedNGDB can reach more than 90\% MRR compared to the central model. 


\subsection{Local Influence}
Because the in-graph queries can be processed by a single local neural graph database, in this part, we evaluate the performance of FedNGDB on these queries to assess the influence of FedNGDB on local queries. We conduct experiments on FB15k-237 and the number of clients is 5. We evaluate the performance of FedNGDB based on GQE and compare the model with no collaboration. The results are summarized in the Figure~\ref{fig:query_types_exp_2}. As shown in the figure, although each client has different performance due to the triplets correlation being different in each sub-dataset, FedNGDB can improve all clients' performance compared to the model without collaboration.


\section{Conclusion \label{sec:conclude}}

In this work, we present a federated neural graph database, FedNGDB to reason over distributed knowledge sets with privacy preservation, allowing graph database holders to collaboratively build a distributed graph reasoning system without sharing raw data. We define the distributed graph complex logical query answering problem. To solve the problem, we propose secret aggregation for federated learning where the aggregated parameters can be kept secret to the server. Besides, we design a distributed query retrieval process for answering queries from distributed graph database sets to protect clients' privacy. To evaluate FedNGDB model performance, we construct a benchmark based on three commonly used knowledge graph complex query answering datasets: FB15k-237, FB15k, and NELL995. Extensive experiments on the benchmark demonstrate the effectiveness of our proposed FedNGDB. FedNGDB can retrieve answers given a query while keeping the sensitive information secret at local graph databases. In the future, we aim to propose new methods for better answering complex queries by exploiting intrinsic information in the distributed neural graph databases. 

\section*{Acknowledgement}
The authors of this paper were supported by the ITSP Platform Research Project (ITS/189/23FP) from ITC of Hong Kong, SAR, China, and the AoE (AoE/E-601/24-N), the RIF (R6021-20) and the GRF (16205322) from RGC of Hong Kong,SAR, China. 

\bibliography{tmlr}

\begin{thebibliography}{75}
\providecommand{\natexlab}[1]{#1}
\providecommand{\url}[1]{\texttt{#1}}
\expandafter\ifx\csname urlstyle\endcsname\relax
  \providecommand{\doi}[1]{doi: #1}\else
  \providecommand{\doi}{doi: \begingroup \urlstyle{rm}\Url}\fi

\bibitem[Alivanistos et~al.(2021)Alivanistos, Berrendorf, Cochez, and Galkin]{alivanistos2021query}
Dimitrios Alivanistos, Max Berrendorf, Michael Cochez, and Mikhail Galkin.
\newblock Query embedding on hyper-relational knowledge graphs.
\newblock In \emph{International Conference on Learning Representations}, 2021.

\bibitem[Arakelyan et~al.(2020)Arakelyan, Daza, Minervini, and Cochez]{arakelyan2020complex}
Erik Arakelyan, Daniel Daza, Pasquale Minervini, and Michael Cochez.
\newblock Complex query answering with neural link predictors.
\newblock In \emph{International Conference on Learning Representations}, 2020.

\bibitem[Bai et~al.()Bai, Zheng, and Song]{baisequential}
Jiaxin Bai, Tianshi Zheng, and Yangqiu Song.
\newblock Sequential query encoding for complex query answering on knowledge graphs.
\newblock \emph{Transactions on Machine Learning Research}.

\bibitem[Bai et~al.(2022)Bai, Wang, Zhang, and Song]{bai2022query2particles}
Jiaxin Bai, Zihao Wang, Hongming Zhang, and Yangqiu Song.
\newblock Query2particles: Knowledge graph reasoning with particle embeddings.
\newblock \emph{Findings of the Association for Computational Linguistics: NAACL 2022-Findings}, 2022.

\bibitem[Bai et~al.(2023{\natexlab{a}})Bai, Liu, Wang, Luo, and Song]{bai2023complex}
Jiaxin Bai, Xin Liu, Weiqi Wang, Chen Luo, and Yangqiu Song.
\newblock Complex query answering on eventuality knowledge graph with implicit logical constraints.
\newblock \emph{Advances in Neural Information Processing Systems}, 36:\penalty0 30534--30553, 2023{\natexlab{a}}.

\bibitem[Bai et~al.(2023{\natexlab{b}})Bai, Luo, Li, Yin, Yin, and Song]{bai2023knowledge}
Jiaxin Bai, Chen Luo, Zheng Li, Qingyu Yin, Bing Yin, and Yangqiu Song.
\newblock Knowledge graph reasoning over entities and numerical values.
\newblock In \emph{Proceedings of the 29th ACM SIGKDD Conference on Knowledge Discovery and Data Mining}, pp.\  57--68, 2023{\natexlab{b}}.

\bibitem[Bai et~al.(2025)Bai, Wang, Zhou, Yin, Fei, Hu, Deng, Cheng, Zheng, Tsang, et~al.]{bai2025top}
Jiaxin Bai, Zihao Wang, Yukun Zhou, Hang Yin, Weizhi Fei, Qi~Hu, Zheye Deng, Jiayang Cheng, Tianshi Zheng, Hong~Ting Tsang, et~al.
\newblock Top ten challenges towards agentic neural graph databases.
\newblock \emph{arXiv preprint arXiv:2501.14224}, 2025.

\bibitem[Bater et~al.(2017)Bater, Elliott, Eggen, Goel, Kho, and Rogers]{bater2017smcql}
Johes Bater, Gregory Elliott, Craig Eggen, Satyender Goel, Abel Kho, and Jennie Rogers.
\newblock Smcql: secure querying for federated databases.
\newblock \emph{Proceedings of the VLDB Endowment}, 10\penalty0 (6):\penalty0 673--684, 2017.

\bibitem[Berger \& Schrefl(2008)Berger and Schrefl]{berger2008federated}
Stefan Berger and Michael Schrefl.
\newblock From federated databases to a federated data warehouse system.
\newblock In \emph{Proceedings of the 41st Annual Hawaii International Conference on System Sciences (HICSS 2008)}, pp.\  394--394. IEEE, 2008.

\bibitem[Besta et~al.(2022)Besta, Iff, Scheidl, Osawa, Dryden, Podstawski, Chen, and Hoefler]{besta2022neural}
Maciej Besta, Patrick Iff, Florian Scheidl, Kazuki Osawa, Nikoli Dryden, Michal Podstawski, Tiancheng Chen, and Torsten Hoefler.
\newblock Neural graph databases.
\newblock In \emph{Learning on Graphs Conference}, pp.\  31--1. PMLR, 2022.

\bibitem[Bollacker et~al.(2008)Bollacker, Evans, Paritosh, Sturge, and Taylor]{bollacker2008freebase}
Kurt Bollacker, Colin Evans, Praveen Paritosh, Tim Sturge, and Jamie Taylor.
\newblock Freebase: a collaboratively created graph database for structuring human knowledge.
\newblock In \emph{Proceedings of the 2008 ACM SIGMOD international conference on Management of data}, pp.\  1247--1250, 2008.

\bibitem[Bordes et~al.(2013)Bordes, Usunier, Garcia-Duran, Weston, and Yakhnenko]{bordes2013translating}
Antoine Bordes, Nicolas Usunier, Alberto Garcia-Duran, Jason Weston, and Oksana Yakhnenko.
\newblock Translating embeddings for modeling multi-relational data.
\newblock \emph{Advances in neural information processing systems}, 26, 2013.

\bibitem[Byrd \& Polychroniadou(2020)Byrd and Polychroniadou]{byrd2020differentially}
David Byrd and Antigoni Polychroniadou.
\newblock Differentially private secure multi-party computation for federated learning in financial applications.
\newblock In \emph{Proceedings of the First ACM International Conference on AI in Finance}, pp.\  1--9, 2020.

\bibitem[Cao et~al.(2019)Cao, Wang, He, Hu, and Chua]{cao2019unifying}
Yixin Cao, Xiang Wang, Xiangnan He, Zikun Hu, and Tat-Seng Chua.
\newblock Unifying knowledge graph learning and recommendation: Towards a better understanding of user preferences.
\newblock In \emph{The world wide web conference}, pp.\  151--161, 2019.

\bibitem[Carlson et~al.(2010)Carlson, Betteridge, Kisiel, Settles, Hruschka, and Mitchell]{carlson2010toward}
Andrew Carlson, Justin Betteridge, Bryan Kisiel, Burr Settles, Estevam Hruschka, and Tom Mitchell.
\newblock Toward an architecture for never-ending language learning.
\newblock In \emph{Proceedings of the AAAI conference on artificial intelligence}, volume~24, pp.\  1306--1313, 2010.

\bibitem[Chen et~al.(2021)Chen, Zhang, Yuan, Jia, and Chen]{chen2021fede}
Mingyang Chen, Wen Zhang, Zonggang Yuan, Yantao Jia, and Huajun Chen.
\newblock Fede: Embedding knowledge graphs in federated setting.
\newblock In \emph{Proceedings of the 10th International Joint Conference on Knowledge Graphs}, pp.\  80--88, 2021.

\bibitem[Chen et~al.(2022{\natexlab{a}})Chen, Zhang, Yao, Chen, Ding, Huang, and Chen]{MaKEr}
Mingyang Chen, Wen Zhang, Zhen Yao, Xiangnan Chen, Mengxiao Ding, Fei Huang, and Huajun Chen.
\newblock Meta-learning based knowledge extrapolation for knowledge graphs in the federated setting.
\newblock In Lud~De Raedt (ed.), \emph{Proceedings of the Thirty-First International Joint Conference on Artificial Intelligence, {IJCAI-22}}, pp.\  1966--1972. International Joint Conferences on Artificial Intelligence Organization, 7 2022{\natexlab{a}}.
\newblock \doi{10.24963/ijcai.2022/273}.
\newblock URL \url{https://doi.org/10.24963/ijcai.2022/273}.
\newblock Main Track.

\bibitem[Chen et~al.(2022{\natexlab{b}})Chen, Zhang, Yuan, Jia, and Chen]{CHEN2022109459}
Mingyang Chen, Wen Zhang, Zonggang Yuan, Yantao Jia, and Huajun Chen.
\newblock Federated knowledge graph completion via embedding-contrastive learning.
\newblock \emph{Knowledge-Based Systems}, 252:\penalty0 109459, 2022{\natexlab{b}}.
\newblock ISSN 0950-7051.
\newblock \doi{https://doi.org/10.1016/j.knosys.2022.109459}.
\newblock URL \url{https://www.sciencedirect.com/science/article/pii/S0950705122007316}.

\bibitem[Chen et~al.(2022{\natexlab{c}})Chen, Zhang, Zhu, Zhou, Yuan, Xu, and Chen]{10.1145/3477495.3531757}
Mingyang Chen, Wen Zhang, Yushan Zhu, Hongting Zhou, Zonggang Yuan, Changliang Xu, and Huajun Chen.
\newblock Meta-knowledge transfer for inductive knowledge graph embedding.
\newblock In \emph{Proceedings of the 45th International ACM SIGIR Conference on Research and Development in Information Retrieval}, SIGIR '22, pp.\  927–937, New York, NY, USA, 2022{\natexlab{c}}. Association for Computing Machinery.
\newblock ISBN 9781450387323.
\newblock \doi{10.1145/3477495.3531757}.
\newblock URL \url{https://doi.org/10.1145/3477495.3531757}.

\bibitem[Choudhary et~al.(2021)Choudhary, Rao, Katariya, Subbian, and Reddy]{choudhary2021probabilistic}
Nurendra Choudhary, Nikhil Rao, Sumeet Katariya, Karthik Subbian, and Chandan Reddy.
\newblock Probabilistic entity representation model for reasoning over knowledge graphs.
\newblock \emph{Advances in Neural Information Processing Systems}, 34:\penalty0 23440--23451, 2021.

\bibitem[Diffie \& Hellman(2022)Diffie and Hellman]{diffie2022new}
Whitfield Diffie and Martin~E Hellman.
\newblock New directions in cryptography.
\newblock In \emph{Democratizing Cryptography: The Work of Whitfield Diffie and Martin Hellman}, pp.\  365--390. 2022.

\bibitem[Duddu et~al.(2020)Duddu, Boutet, and Shejwalkar]{duddu2020quantifying}
Vasisht Duddu, Antoine Boutet, and Virat Shejwalkar.
\newblock Quantifying privacy leakage in graph embedding.
\newblock In \emph{MobiQuitous 2020-17th EAI International Conference on Mobile and Ubiquitous Systems: Computing, Networking and Services}, pp.\  76--85, 2020.

\bibitem[Dwork(2006)]{dwork2006differential}
Cynthia Dwork.
\newblock Differential privacy.
\newblock In \emph{International colloquium on automata, languages, and programming}, pp.\  1--12. Springer, 2006.

\bibitem[Dwork(2008)]{dwork2008differential}
Cynthia Dwork.
\newblock Differential privacy: A survey of results.
\newblock In \emph{International conference on theory and applications of models of computation}, pp.\  1--19. Springer, 2008.

\bibitem[Dwork et~al.(2014)Dwork, Roth, et~al.]{dwork2014algorithmic}
Cynthia Dwork, Aaron Roth, et~al.
\newblock The algorithmic foundations of differential privacy.
\newblock \emph{Foundations and Trends{\textregistered} in Theoretical Computer Science}, 9\penalty0 (3--4):\penalty0 211--407, 2014.

\bibitem[Gal{\'a}rraga et~al.(2013)Gal{\'a}rraga, Teflioudi, Hose, and Suchanek]{galarraga2013amie}
Luis~Antonio Gal{\'a}rraga, Christina Teflioudi, Katja Hose, and Fabian Suchanek.
\newblock Amie: association rule mining under incomplete evidence in ontological knowledge bases.
\newblock In \emph{Proceedings of the 22nd international conference on World Wide Web}, pp.\  413--422, 2013.

\bibitem[Gao et~al.(2023)Gao, Xiong, Gao, Jia, Pan, Bi, Dai, Sun, and Wang]{gao2023retrieval}
Yunfan Gao, Yun Xiong, Xinyu Gao, Kangxiang Jia, Jinliu Pan, Yuxi Bi, Yi~Dai, Jiawei Sun, and Haofen Wang.
\newblock Retrieval-augmented generation for large language models: A survey.
\newblock \emph{arXiv preprint arXiv:2312.10997}, 2023.

\bibitem[Geyer et~al.(2017)Geyer, Klein, and Nabi]{geyer2017differentially}
Robin~C Geyer, Tassilo Klein, and Moin Nabi.
\newblock Differentially private federated learning: A client level perspective.
\newblock \emph{arXiv preprint arXiv:1712.07557}, 2017.

\bibitem[Hamilton et~al.(2018)Hamilton, Bajaj, Zitnik, Jurafsky, and Leskovec]{hamilton2018embedding}
Will Hamilton, Payal Bajaj, Marinka Zitnik, Dan Jurafsky, and Jure Leskovec.
\newblock Embedding logical queries on knowledge graphs.
\newblock \emph{Advances in neural information processing systems}, 31, 2018.

\bibitem[Hu et~al.(2022{\natexlab{a}})Hu, Liu, Zhang, Wang, Lin, and Luo]{hu2022federated}
Ce~Hu, Baisong Liu, Xueyuan Zhang, Zhiye Wang, Chennan Lin, and Linze Luo.
\newblock A federated multi-server knowledge graph embedding framework for link prediction.
\newblock In \emph{2022 IEEE 34th International Conference on Tools with Artificial Intelligence (ICTAI)}, pp.\  366--371. IEEE, 2022{\natexlab{a}}.

\bibitem[Hu et~al.(2022{\natexlab{b}})Hu, Cheng, Vap, and Borowczak]{hu2022learning}
Hui Hu, Lu~Cheng, Jayden~Parker Vap, and Mike Borowczak.
\newblock Learning privacy-preserving graph convolutional network with partially observed sensitive attributes.
\newblock In \emph{Proceedings of the ACM Web Conference 2022}, pp.\  3552--3561, 2022{\natexlab{b}}.

\bibitem[Hu \& Song(2023)Hu and Song]{hu2023independent}
Qi~Hu and Yangqiu Song.
\newblock Independent distribution regularization for private graph embedding.
\newblock In \emph{Proceedings of the 32nd ACM International Conference on Information and Knowledge Management}, CIKM '23, pp.\  823–832, New York, NY, USA, 2023. Association for Computing Machinery.
\newblock ISBN 9798400701245.
\newblock \doi{10.1145/3583780.3614933}.
\newblock URL \url{https://doi.org/10.1145/3583780.3614933}.

\bibitem[Hu \& Song(2024)Hu and Song]{hu2024user}
Qi~Hu and Yangqiu Song.
\newblock User consented federated recommender system against personalized attribute inference attack.
\newblock In \emph{Proceedings of the 17th ACM International Conference on Web Search and Data Mining}, pp.\  276--285, 2024.

\bibitem[Hu et~al.(2024)Hu, Li, Bai, Wang, and Song]{hu2024privacy}
Qi~Hu, Haoran Li, Jiaxin Bai, Zihao Wang, and Yangqiu Song.
\newblock Privacy-preserved neural graph databases.
\newblock In \emph{Proceedings of the 30th ACM SIGKDD Conference on Knowledge Discovery and Data Mining}, pp.\  1108--1118, 2024.

\bibitem[Hu et~al.(2023)Hu, Liang, Wu, Xiao, Wang, Li, Liu, and Qin]{10.1145/3543507.3583450}
Yuke Hu, Wei Liang, Ruofan Wu, Kai Xiao, Weiqiang Wang, Xiaochen Li, Jinfei Liu, and Zhan Qin.
\newblock Quantifying and defending against privacy threats on federated knowledge graph embedding.
\newblock In \emph{Proceedings of the ACM Web Conference 2023}, WWW '23, pp.\  2306–2317, New York, NY, USA, 2023. Association for Computing Machinery.
\newblock ISBN 9781450394161.
\newblock \doi{10.1145/3543507.3583450}.
\newblock URL \url{https://doi.org/10.1145/3543507.3583450}.

\bibitem[Huang et~al.(2022)Huang, Liu, Li, Ji, Wang, and Huang]{huang2022fedcke}
Wei Huang, Jia Liu, Tianrui Li, Shenggong Ji, Dexian Wang, and Tianqiang Huang.
\newblock Fedcke: Cross-domain knowledge graph embedding in federated learning.
\newblock \emph{IEEE Transactions on Big Data}, 2022.

\bibitem[Hussien et~al.(2025)Hussien, Melo, Ballardini, Maldonado, Izquierdo, and Sotelo]{hussien2025rag}
Mohamed~Manzour Hussien, Angie~Nataly Melo, Augusto~Luis Ballardini, Carlota~Salinas Maldonado, Rub{\'e}n Izquierdo, and Miguel~Angel Sotelo.
\newblock Rag-based explainable prediction of road users behaviors for automated driving using knowledge graphs and large language models.
\newblock \emph{Expert Systems with Applications}, 265:\penalty0 125914, 2025.

\bibitem[Jin et~al.(2024)Jin, Yu, Shu, Zhang, Fan, Hua, Zhu, Meng, Wang, Du, et~al.]{jin2024health}
Mingyu Jin, Qinkai Yu, Dong Shu, Chong Zhang, Lizhou Fan, Wenyue Hua, Suiyuan Zhu, Yanda Meng, Zhenting Wang, Mengnan Du, et~al.
\newblock Health-llm: Personalized retrieval-augmented disease prediction system.
\newblock \emph{arXiv preprint arXiv:2402.00746}, 2024.

\bibitem[Kotnis et~al.(2021)Kotnis, Lawrence, and Niepert]{kotnis2021answering}
Bhushan Kotnis, Carolin Lawrence, and Mathias Niepert.
\newblock Answering complex queries in knowledge graphs with bidirectional sequence encoders.
\newblock In \emph{Proceedings of the AAAI Conference on Artificial Intelligence}, volume~35, pp.\  4968--4977, 2021.

\bibitem[Leblay \& Chekol(2018)Leblay and Chekol]{leblay2018deriving}
Julien Leblay and Melisachew~Wudage Chekol.
\newblock Deriving validity time in knowledge graph.
\newblock In \emph{Companion Proceedings of the The Web Conference 2018}, pp.\  1771--1776, 2018.

\bibitem[Lewis et~al.(2020)Lewis, Perez, Piktus, Petroni, Karpukhin, Goyal, K{\"u}ttler, Lewis, Yih, Rockt{\"a}schel, et~al.]{lewis2020retrieval}
Patrick Lewis, Ethan Perez, Aleksandra Piktus, Fabio Petroni, Vladimir Karpukhin, Naman Goyal, Heinrich K{\"u}ttler, Mike Lewis, Wen-tau Yih, Tim Rockt{\"a}schel, et~al.
\newblock Retrieval-augmented generation for knowledge-intensive nlp tasks.
\newblock \emph{Advances in Neural Information Processing Systems}, 33:\penalty0 9459--9474, 2020.

\bibitem[Li(2013)]{10.5555/2519214}
Yingjie Li.
\newblock \emph{A federated query answering system for semantic web data}.
\newblock PhD thesis, USA, 2013.

\bibitem[Liu et~al.(2021)Liu, Du, Ji, Zhai, and Tong]{liu2021neural}
Lihui Liu, Boxin Du, Heng Ji, ChengXiang Zhai, and Hanghang Tong.
\newblock Neural-answering logical queries on knowledge graphs.
\newblock In \emph{Proceedings of the 27th ACM SIGKDD conference on knowledge discovery \& data mining}, pp.\  1087--1097, 2021.

\bibitem[Liu et~al.(2022)Liu, Zhao, Su, Cen, Qiu, Zhang, Wu, Dong, and Tang]{liu2022mask}
Xiao Liu, Shiyu Zhao, Kai Su, Yukuo Cen, Jiezhong Qiu, Mengdi Zhang, Wei Wu, Yuxiao Dong, and Jie Tang.
\newblock Mask and reason: Pre-training knowledge graph transformers for complex logical queries.
\newblock In \emph{Proceedings of the 28th ACM SIGKDD Conference on Knowledge Discovery and Data Mining}, pp.\  1120--1130, 2022.

\bibitem[Long et~al.(2020)Long, Tan, Jiang, and Zhang]{long2020federated}
Guodong Long, Yue Tan, Jing Jiang, and Chengqi Zhang.
\newblock Federated learning for open banking.
\newblock In \emph{Federated Learning: Privacy and Incentive}, pp.\  240--254. Springer, 2020.

\bibitem[Loshchilov \& Hutter(2018)Loshchilov and Hutter]{loshchilov2018decoupled}
Ilya Loshchilov and Frank Hutter.
\newblock Decoupled weight decay regularization.
\newblock In \emph{International Conference on Learning Representations}, 2018.

\bibitem[Matsumoto et~al.(2024)Matsumoto, Moran, Choi, Hernandez, Venkatesan, Wang, and Moore]{matsumoto2024kragen}
Nicholas Matsumoto, Jay Moran, Hyunjun Choi, Miguel~E Hernandez, Mythreye Venkatesan, Paul Wang, and Jason~H Moore.
\newblock Kragen: a knowledge graph-enhanced rag framework for biomedical problem solving using large language models.
\newblock \emph{Bioinformatics}, 40\penalty0 (6), 2024.

\bibitem[McMahan et~al.(2017)McMahan, Moore, Ramage, Hampson, and y~Arcas]{mcmahan2017communication}
Brendan McMahan, Eider Moore, Daniel Ramage, Seth Hampson, and Blaise~Aguera y~Arcas.
\newblock Communication-efficient learning of deep networks from decentralized data.
\newblock In \emph{Artificial Intelligence and Statistics}, pp.\  1273--1282. PMLR, 2017.

\bibitem[Mohassel \& Zhang(2017)Mohassel and Zhang]{mohassel2017secureml}
Payman Mohassel and Yupeng Zhang.
\newblock Secureml: A system for scalable privacy-preserving machine learning.
\newblock In \emph{2017 IEEE symposium on security and privacy (SP)}, pp.\  19--38. IEEE, 2017.

\bibitem[Nadal et~al.(2021)Nadal, Abell{\'o}, Romero, Vansummeren, and Vassiliadis]{nadal2021graph}
Sergi Nadal, Alberto Abell{\'o}, Oscar Romero, Stijn Vansummeren, and Panos Vassiliadis.
\newblock Graph-driven federated data management.
\newblock \emph{IEEE Transactions on Knowledge and Data Engineering}, 35\penalty0 (1):\penalty0 509--520, 2021.

\bibitem[Paillier(1999)]{paillier1999public}
Pascal Paillier.
\newblock Public-key cryptosystems based on composite degree residuosity classes.
\newblock In \emph{International conference on the theory and applications of cryptographic techniques}, pp.\  223--238. Springer, 1999.

\bibitem[Peng et~al.(2021)Peng, Li, Song, Zheng, and Li]{peng2021differentially}
Hao Peng, Haoran Li, Yangqiu Song, Vincent Zheng, and Jianxin Li.
\newblock Differentially private federated knowledge graphs embedding.
\newblock In \emph{Proceedings of the 30th ACM International Conference on Information \& Knowledge Management}, pp.\  1416--1425, 2021.

\bibitem[Prusti et~al.(2021)Prusti, Das, and Rath]{prusti2021credit}
Debachudamani Prusti, Daisy Das, and Santanu~Kumar Rath.
\newblock Credit card fraud detection technique by applying graph database model.
\newblock \emph{Arabian Journal for Science and Engineering}, 46\penalty0 (9):\penalty0 1--20, 2021.

\bibitem[Ren et~al.(2020)Ren, Hu, and Leskovec]{ren2020query2box}
H~Ren, W~Hu, and J~Leskovec.
\newblock Query2box: Reasoning over knowledge graphs in vector space using box embeddings.
\newblock In \emph{International Conference on Learning Representations (ICLR)}, 2020.

\bibitem[Ren \& Leskovec(2020)Ren and Leskovec]{ren2020beta}
Hongyu Ren and Jure Leskovec.
\newblock Beta embeddings for multi-hop logical reasoning in knowledge graphs.
\newblock \emph{Advances in Neural Information Processing Systems}, 33:\penalty0 19716--19726, 2020.

\bibitem[Ren et~al.(2023)Ren, Galkin, Cochez, Zhu, and Leskovec]{ren2023neural}
Hongyu Ren, Mikhail Galkin, Michael Cochez, Zhaocheng Zhu, and Jure Leskovec.
\newblock Neural graph reasoning: Complex logical query answering meets graph databases.
\newblock \emph{arXiv preprint arXiv:2303.14617}, 2023.

\bibitem[Sadowski \& Rathle(2014)Sadowski and Rathle]{sadowski2014fraud}
Gorka Sadowski and Philip Rathle.
\newblock Fraud detection: Discovering connections with graph databases.
\newblock \emph{White Paper-Neo Technology-Graphs are Everywhere}, 13:\penalty0 1--13, 2014.

\bibitem[Tang et~al.(2023)Tang, Wu, Cao, Liao, and Zhou]{tang2023fedmkgc}
Wei Tang, Zhiqian Wu, Yixin Cao, Yong Liao, and Pengyuan Zhou.
\newblock Fedmkgc: Privacy-preserving federated multilingual knowledge graph completion.
\newblock \emph{arXiv preprint arXiv:2312.10645}, 2023.

\bibitem[Toutanova \& Chen(2015)Toutanova and Chen]{toutanova2015observed}
Kristina Toutanova and Danqi Chen.
\newblock Observed versus latent features for knowledge base and text inference.
\newblock In \emph{Proceedings of the 3rd workshop on continuous vector space models and their compositionality}, pp.\  57--66, 2015.

\bibitem[Wang et~al.(2023{\natexlab{a}})Wang, Chen, and Grau]{wang2023efficient}
Dingmin Wang, Yeyuan Chen, and Bernardo~Cuenca Grau.
\newblock Efficient embeddings of logical variables for query answering over incomplete knowledge graphs.
\newblock In \emph{Proceedings of the AAAI Conference on Artificial Intelligence}, volume~37, pp.\  4652--4659, 2023{\natexlab{a}}.

\bibitem[Wang et~al.(2023{\natexlab{b}})Wang, Zeng, Xu, Guo, and Zhao]{wang2023federated}
Maolin Wang, Dun Zeng, Zenglin Xu, Ruocheng Guo, and Xiangyu Zhao.
\newblock Federated knowledge graph completion via latent embedding sharing and tensor factorization.
\newblock In \emph{2023 IEEE International Conference on Data Mining (ICDM)}, pp.\  1361--1366. IEEE, 2023{\natexlab{b}}.

\bibitem[Wang et~al.(2015)Wang, Wang, and Guo]{wang2015knowledge}
Quan Wang, Bin Wang, and Li~Guo.
\newblock Knowledge base completion using embeddings and rules.
\newblock In \emph{Twenty-fourth international joint conference on artificial intelligence}, 2015.

\bibitem[Wang et~al.(2019)Wang, He, Cao, Liu, and Chua]{wang2019kgat}
Xiang Wang, Xiangnan He, Yixin Cao, Meng Liu, and Tat-Seng Chua.
\newblock Kgat: Knowledge graph attention network for recommendation.
\newblock In \emph{Proceedings of the 25th ACM SIGKDD international conference on knowledge discovery \& data mining}, pp.\  950--958, 2019.

\bibitem[Wang et~al.(2022)Wang, Song, Wong, and See]{wang2022logical}
Zihao Wang, Yangqiu Song, Ginny Wong, and Simon See.
\newblock Logical message passing networks with one-hop inference on atomic formulas.
\newblock In \emph{The Eleventh International Conference on Learning Representations}, 2022.

\bibitem[Xi et~al.(2022)Xi, Pang, Li, Du, Ji, Ma, and Wang]{xi2022reasoning}
Zhaohan Xi, Ren Pang, Changjiang Li, Tianyu Du, Shouling Ji, Fenglong Ma, and Ting Wang.
\newblock Reasoning over multi-view knowledge graphs.
\newblock \emph{arXiv preprint arXiv:2209.13702}, 2022.

\bibitem[Yang et~al.(2022)Yang, Qing, Li, Lu, and Lin]{yang2022gammae}
Dong Yang, Peijun Qing, Yang Li, Haonan Lu, and Xiaodong Lin.
\newblock Gammae: Gamma embeddings for logical queries on knowledge graphs.
\newblock In \emph{Proceedings of the 2022 Conference on Empirical Methods in Natural Language Processing}, pp.\  745--760, 2022.

\bibitem[Yang et~al.(2020)Yang, Tan, Zheng, Chen, and Yang]{yang2020federated}
Liu Yang, Ben Tan, Vincent~W Zheng, Kai Chen, and Qiang Yang.
\newblock Federated recommendation systems.
\newblock \emph{Federated Learning: Privacy and Incentive}, pp.\  225--239, 2020.

\bibitem[Yang et~al.(2019)Yang, Liu, Chen, and Tong]{yang2019federated}
Qiang Yang, Yang Liu, Tianjian Chen, and Yongxin Tong.
\newblock Federated machine learning: Concept and applications.
\newblock \emph{ACM Transactions on Intelligent Systems and Technology (TIST)}, 10\penalty0 (2):\penalty0 1--19, 2019.

\bibitem[Yi et~al.(2021)Yi, Wu, Wu, Liu, Sun, and Xie]{yi2021efficient}
Jingwei Yi, Fangzhao Wu, Chuhan Wu, Ruixuan Liu, Guangzhong Sun, and Xing Xie.
\newblock Efficient-fedrec: Efficient federated learning framework for privacy-preserving news recommendation.
\newblock In \emph{Proceedings of the 2021 Conference on Empirical Methods in Natural Language Processing}, pp.\  2814--2824, 2021.

\bibitem[Zhang et~al.(2020)Zhang, Li, Xia, Wang, Yan, and Liu]{zhang2020batchcrypt}
Chengliang Zhang, Suyi Li, Junzhe Xia, Wei Wang, Feng Yan, and Yang Liu.
\newblock $\{$BatchCrypt$\}$: Efficient homomorphic encryption for $\{$Cross-Silo$\}$ federated learning.
\newblock In \emph{2020 USENIX annual technical conference (USENIX ATC 20)}, pp.\  493--506, 2020.

\bibitem[Zhang et~al.(2024)Zhang, Yang, Ying, and Lauw]{zhang2024text}
Delvin~Ce Zhang, Menglin Yang, Rex Ying, and Hady~W Lauw.
\newblock Text-attributed graph representation learning: Methods, applications, and challenges.
\newblock In \emph{Companion Proceedings of the ACM Web Conference 2024}, pp.\  1298--1301, 2024.

\bibitem[Zhang et~al.(2022)Zhang, Wang, Wang, Huang, Yang, Chen, and Sun]{zhang2022efficient}
Kai Zhang, Yu~Wang, Hongyi Wang, Lifu Huang, Carl Yang, Xun Chen, and Lichao Sun.
\newblock Efficient federated learning on knowledge graphs via privacy-preserving relation embedding aggregation.
\newblock In \emph{Findings of the Association for Computational Linguistics: EMNLP 2022}, pp.\  613--621, 2022.

\bibitem[Zhang et~al.(2021)Zhang, Wang, Chen, Ji, and Wu]{zhang2021cone}
Zhanqiu Zhang, Jie Wang, Jiajun Chen, Shuiwang Ji, and Feng Wu.
\newblock Cone: Cone embeddings for multi-hop reasoning over knowledge graphs.
\newblock \emph{Advances in Neural Information Processing Systems}, 34:\penalty0 19172--19183, 2021.

\bibitem[Zhu et~al.(2019)Zhu, Liu, and Han]{zhu2019deep}
Ligeng Zhu, Zhijian Liu, and Song Han.
\newblock Deep leakage from gradients.
\newblock \emph{Advances in neural information processing systems}, 32, 2019.

\bibitem[Zhu et~al.(2022)Zhu, Galkin, Zhang, and Tang]{zhu2022neural}
Zhaocheng Zhu, Mikhail Galkin, Zuobai Zhang, and Jian Tang.
\newblock Neural-symbolic models for logical queries on knowledge graphs.
\newblock In \emph{International Conference on Machine Learning}, pp.\  27454--27478. PMLR, 2022.

\end{thebibliography}
\bibliographystyle{tmlr}

\newpage

\appendix

\section{Parameter Sharing \label{appendix:sharing}}

In this section, we provide an example of sharing secrets between clients. The parameters can be shared under the protection using various encryption methods, for example, the commonly used is Diffie–Hellman key exchange~\citep{diffie2022new} shown as follows, we consider the sharing process between two clients:  
\begin{itemize}
    \item Client A and Client B publicly agree to use a modulus $p$ and base $g$, $p$ is a prime.
    \item Client A chooses a secret integer $a$, then sends Client B $m_A = g^a \mod p$.
    \item Client B chooses a secret integer $b$, then sends Client A $m_B = g^b \mod p$.
    \item Client A computes $s = m_B^a \mod p$.
    \item Client B computes $s = m_A^b \mod p$.
\end{itemize}
After D-H key exchange, Client A and B share a secret $s = g^{ab} \mod p$. The secret $s$ can be used as encryption to share sensitive information between clients.

\section{Alogrithm}
\subsection{Secret Aggregation}
We present the pseudo-code of secret aggregation in Algorithm \ref{alg:aggregation}. 
\begin{algorithm}[ht]
\caption{Secret Aggregation}
\label{alg:aggregation}
\begin{algorithmic}
\REQUIRE $n$ clients $C_1, C_2, \ldots, C_n$, client $C_i$ has parameter $\theta_i$ 

\STATE{Each client $C_i$ generates random parameter $\theta_{i}^r$}
\STATE Transmit $\theta_i^r$ to all other clients with encryption, each\\ client has a set of parameters $\{\theta_{1}^r,\theta_{2}^r.\cdots,\theta_{n}^r\}$

\noindent\STATE {\bf Client $C_i$:}
    \STATE{Upload perturbed parameter $(\theta_i+\theta_i^r)$ to server with HE Encryption.}
    \STATE{Receive encrypted parameters $\sum_{j=1}^n (\theta_j + \theta_j^r)$} from server and HE decryption.
    \STATE{Compute encrypted parameters $\theta = [\sum_{j=1}^n (\theta_j + \theta_j^r) - \sum_i^n \theta_i^r ]/n$}
    
\noindent\STATE {\bf Server:}
    \FOR{$i=1,\cdots,n$}
        \STATE{Receive encrypted parameters $(\theta_i+\theta_i^r)$ from client $C_i$}
    \ENDFOR
    \STATE{Compute encrypted parameters $\theta^r=\sum_{j=1}^n (\theta_j + \theta_j^r)$}
    \STATE{Send encrypted parameters $\theta^r$ to all clients}
\end{algorithmic}
\end{algorithm}

\begin{algorithm}[ht]
\caption{FedNGDB Framework}
\label{alg:FedNGDB}
\begin{algorithmic}[0]
\REQUIRE 
The number of clients $N$; The faction of clients\\ selected in each round $F$; 

\noindent\STATE {\bf Client $C_i$:}
    \STATE{Client $C_i$ initialize entity embeddings $\mathbf{E}^i$ and perturbation embeddings $\mathbf{E}_r^i$.}
    \STATE{Share $\mathbf{E}_r^i$ with other clients with encryption}
    \STATE{Receive and decryption to get $\{\mathbf{E}_r^1,\cdots,\mathbf{E}_r^N\}$ }
    \STATE{Upload $\mathbf{E}^i$ to server for secret aggregation, receive $\mathbf{E}_0^i$}

\noindent\STATE {\bf Server:}
\STATE {Server constructs permutation matrices $\{M_i\}_{i=1}^N$, and existence vectors $\{v_i\}_{i=1}^N$ and initialize operator networks $\Theta_0$, distribute to all clients.}
\FOR{$t=0,1,2,\cdots$}
    \STATE Server distributes operator networks to each client.
    \STATE $\mathcal{C}_t$ $\leftarrow$ Randomly select client set with $N\times F$ clients.
    \FOR{$C_i \in \mathcal{C}_t$ \textbf{in parallel}}
        \STATE $(\mathbf{E}^i_r+\mathbf{E}^i_{t+1}), \Theta^{i}_{t+1} \leftarrow \textbf{ClientUpdate}(C_i, {\mathbf{M}^i}^\top \mathbf{E}^r_t, \Theta_t)$
    \ENDFOR
\STATE $    \mathbf{E}_{t+1}^r \leftarrow \left( \mathbbm{1} \oslash \sum_{i \in \mathbf{C}_t} \mathbf{v}^{i} \right) \otimes \sum_{i \in \mathbf{C}_t}\mathbf{M}^{i} (\mathbf{E}_{t+1}^i+\mathbf{E}_r^i)$
\STATE $\Theta_{t+1} \leftarrow 1/|\mathcal{C}_t| \sum_{i \in \mathcal{C}_t} \Theta_t^i$
\ENDFOR
    
\noindent\STATE {\bf ClientUpdate$(C_i, \mathbf{E}^r, \Theta$):}
    \STATE $\mathbf{E} \leftarrow  \mathbf{E}^{r} 
 - {\mathbf{M}^i}^\top \left( \mathbbm{1} \oslash \sum_{j \in \mathbf{C}} \mathbf{v}^{j} \right) \otimes \sum_{j \in \mathbf{C}}\mathbf{M}^{j} \mathbf{E}^{j}_{r}$
    \FOR{$e=1,\cdots,E$}
        \STATE{$\mathbf{E}, \Theta \leftarrow \textbf{LocalUpdate}(\mathbf{E}, \Theta)$}
    \ENDFOR
    \RETURN $\mathbf{E}+\mathbf{E}^i_r, \Theta$
\end{algorithmic}
\end{algorithm}

\subsection{FedNGDB Framework}
We present the pseudo-code of FedNGDB in Algorithm \ref{alg:FedNGDB}.

\subsection{Differential Privacy Setting \label{sec:DP}}
Differential privacy represents a mathematical framework for quantifying and limiting the disclosure of private information in statistical databases~\citep{dwork2008differential}. In federated learning, differential privacy enables collaborative model training across decentralized data sources while providing formal privacy guarantees. By injecting calibrated noise into the training process, these techniques ensure that individual contributions cannot be reliably identified or extracted from the global model~\citep{geyer2017differentially}. The privacy-utility tradeoff can be tuned via the privacy budget parameter $\epsilon$, allowing practitioners to balance learning performance against disclosure risk. The formal definition of $\epsilon$-differential privacy ($\epsilon$-DP) is as follows~\citep{dwork2006differential}:
A randomized algorithm $M$ satisfies $\epsilon$-differential privacy if for all neighboring datasets $D_1$ and $D_2$ that differ in at most one record, and for all possible outputs $S \subseteq Range(M)$:
$$
Pr[M(D_1) \in S] \leq e^\epsilon Pr[M(D_2) \in S]
$$
Where $\epsilon$ is the privacy parameter that quantifies the privacy loss. 

In the former part, we introduce Laplacian noise to the local gradients to meet the guarantee of $\epsilon$-DP for in-graph query in local neural graph databases. However, for cross-graph query, we compute the score on each graph database for each query. The Post-Processing Theorem~\citep{dwork2014algorithmic} which states the composition of a data-independent mapping with an $\epsilon$-DP algorithm $M$ is also $\epsilon$-DP. Since the ranking operation of FedNGDB is a post-processing step applied to the outputs $(M_1(D),\cdots,M_n(D))$ from each local databases and the overlapped candidate answers share same scores, by the post-processing theorem of differential privacy, the final ranked list maintains the same privacy guarantee of $\epsilon$-DP.

\section{Auxiliary Experiments}
Here we present some auxiliary experiments to further evaluate the performance of FedNGDB.

\subsection{More Clients \label{appx: more clients}}
In the former experiments, we evaluate FedNGDB's performance when there are 3 or 5 clients in the federated system. To further evaluate models' performance when there are more clients. We split the graph-structured data into 10 subgraphs and evaluate the retrieval performance of FedNGDB using GQE as the base model. As shown in Table \ref{tab:more}, FedNGDB can still improve the retrieval performance compared to local training when there are more clients participating in the distributed system, demonstrating the effectiveness of FedNGDB.

\begin{table}[ht]
\centering
\caption{The performance of GQE when \#C=10.}
\label{tab:more}
\scalebox{0.79}{
\begin{tabular}{c|c|rrrr|rrrr|rrrr}

\toprule

                            & \multicolumn{1}{c|}{}                          & \multicolumn{4}{c|}{FB15k-237}                                              & \multicolumn{4}{c|}{FB15k}                                              & \multicolumn{4}{c}{NELL995}                                           \\ \cline{3-14} 
                            & \multicolumn{1}{c|}{}                          & \multicolumn{2}{c}{In-graph}      & \multicolumn{2}{c|}{Cross-graph}  & \multicolumn{2}{c}{In-graph}      & \multicolumn{2}{c|}{Cross-graph}  & \multicolumn{2}{c}{In-graph}       & \multicolumn{2}{c}{Cross-graph}    \\
\multirow{-3}{*}{Graph}     & \multirow{-3}{*}{Setting} & HR@3 & MRR & HR@3 & MRR & HR@3 & MRR & HR@3 & MRR & HR@3 & MRR  & HR@3 & MRR  \\ \midrule
                            & Local  &8.96&8.47& - & - & 13.46 & 13.04 &-  & - &8.36  &7.83 & - &- \\
                            & Central&13.15 & 12.76 & 13.24 & 12.89 & 19.76&19.43&20.27 & 19.87 & 13.76 & 13.28 & 13.53 & 13.27 \\
\multirow{-3}{*}{GQE} & FedNGDB  &10.67 & 10.17 & 10.23 & 10.04 & 15.39 & 14.98& 14.86 & 14.45 & 12.93 & 12.11 & 12.55 & 11.96\\
\bottomrule
\end{tabular}}
\end{table}  

\subsection{Relation Overlap}
In the experiments, we evaluate the retrieval performance when there are no overlap relations between local graph databases, however, various graph databases can have shared relations in real life, to evaluate the performance in such a scenario, we evaluate the FedNGDB with GQE's retrieval performance. The graph-structured data is randomly split into 3 subgraphs. The results are shown in Table \ref{tab:overlap}, showing that FedNGDB can successfully retrieve answers from distributed graph databases. 

\begin{table}[ht]
\caption{The MRR of GQE when relation overlapped.}
\centering
\label{tab:overlap}
\scalebox{0.79}{
\begin{tabular}{c|c|rrrr|rrrr|rrrr}

\toprule

                            & \multicolumn{1}{c|}{}                          & \multicolumn{4}{c|}{FB15k-237}                                              & \multicolumn{4}{c|}{FB15k}                                              & \multicolumn{4}{c}{NELL995}                                           \\ \cline{3-14} 
                            & \multicolumn{1}{c|}{}                          & \multicolumn{2}{c}{In-graph}      & \multicolumn{2}{c|}{Cross-graph}  & \multicolumn{2}{c}{In-graph}      & \multicolumn{2}{c|}{Cross-graph}  & \multicolumn{2}{c}{In-graph}       & \multicolumn{2}{c}{Cross-graph}    \\
\multirow{-3}{*}{Graph}     & \multirow{-3}{*}{Setting} & HR@3 & MRR & HR@3 & MRR & HR@3 & MRR & HR@3 & MRR & HR@3 & MRR  & HR@3 & MRR  \\ \midrule
                            & Local  &10.48&10.22& - & - & 20.76 & 20.21 &-  & - &10.03  &9.64 & - &- \\
                            & Central&13.47 & 13.27 & 13.86 & 13.54 & 30.74&30.46&30.27 & 30.16 & 13.82 & 13.68 & 13.56 & 13.12 \\
\multirow{-3}{*}{GQE} & FedNGDB  &11.96 & 11.42 & 11.67 & 11.43 & 23.89 & 22.47& 23.22 & 22.89 & 11.68 & 11.36 & 12.03 & 11.89\\
\bottomrule
\end{tabular}}
\end{table}

\subsection{Convergence Rate}
We evaluate the convergence speed of three federated frameworks. The results are presented by the average number of communication round ratios relative to FedE. As shown in Table \ref{tab:convergence}, FedNGDB's convergence speed is faster than FedR while slightly slower than FedE, demonstrating that our FedNGDB can protect stronger protection while remaining competitive efficiency.

\begin{table}[h]
\caption{The statistics of communication rounds .}
\centering
\label{tab:convergence}
\begin{tabular}{cccccc}
\toprule
Setting    & FedE & FedR   & FedNGDB \\
\midrule
Relative Rounds to FedE & 1.00 &1.32  & 1.09 \\

\bottomrule
\end{tabular}
\end{table}

\end{document}